%% file: manuscript.tex
\documentclass[runningheads]{llncs}

\usepackage{eccv}

\makeatletter
\let\ifreview\ifeccv@review
\makeatother

\usepackage[accsupp]{axessibility}  
\usepackage{amsmath}
\usepackage{amssymb}
\usepackage{array}
\usepackage{booktabs}
\usepackage{graphicx}
\usepackage[misc]{ifsym}
\usepackage{ifthen}
\usepackage{makecell}
\usepackage{mathtools}
\usepackage{multirow}
\usepackage{pifont}
\usepackage{tabularx}
\usepackage{xspace}

\usepackage[pagebackref,breaklinks,colorlinks,citecolor=blue,linkcolor=red,bookmarks=false]{hyperref}
\usepackage{orcidlink}

\newcolumntype{Y}{>{\centering\arraybackslash}X}
\newcommand{\OM}{MoTok\xspace}

\newcommand{\eg}{\textit{e.g.}}

\newcommand{\darr}{$\downarrow$\xspace}
\newcommand{\uarr}{$\uparrow$\xspace}
\newcommand{\rarr}{$\rightarrow$\xspace}
\newcommand{\cmark}{\textcolor{green}{\ding{51}}}
\newcommand{\xmark}{\textcolor{red}{\ding{55}}}
\newcommand{\std}[3][]{%
  \ifthenelse{\equal{#1}{1}}{%
    \textcolor{red}{$\mathbf{#2}^{\pm #3}$}%
  }{%
    \ifthenelse{\equal{#1}{2}}{%
      \textcolor{blue}{$\mathbf{#2}^{\pm #3}$}%
    }{%
      $#2^{\pm #3}$%
    }%
  }
}
\newcommand{\mean}[2][]{%
  \ifthenelse{\equal{#1}{1}}{%
    \textcolor{red}{$\mathbf{#2}$}%
  }{%
    \ifthenelse{\equal{#1}{2}}{%
      \textcolor{blue}{$\mathbf{#2}$}%
    }{%
      $#2$%
    }%
  }
}
\renewcommand{\paragraph}[1]{%
  \par\medskip
  \noindent\textbf{#1.\ }%
}

\begin{document}

\title{Bridging Semantic and Kinematic Conditions \\ 
with Diffusion-based Discrete Motion Tokenizer} 
\titlerunning{\OM: Diffusion-based Discrete Motion Tokenizer}

\author{
Chenyang Gu\textsuperscript{1,*}~\orcidlink{0009-0009-4536-5437} \and
Mingyuan Zhang\textsuperscript{1,*}~\orcidlink{0000-0001-8212-715X} \and
Haozhe Xie\textsuperscript{1,*}~\orcidlink{0000-0001-9596-5179} \and\\
Zhongang Cai\textsuperscript{1}~\orcidlink{0000-0002-1810-3855} \and
Lei Yang\textsuperscript{2}~\orcidlink{0000-0002-0571-5924} \and
Ziwei Liu\textsuperscript{1, \Letter}~\orcidlink{0000-0002-4220-5958}}

\authorrunning{C. Gu et al.}

\institute{%
\textsuperscript{1}S-Lab, Nanyang Technological University \hspace{2 mm}
\textsuperscript{2}The Chinese University of Hong Kong\\
\url{https://rheallyc.github.io/projects/motok}}

\maketitle
\ifreview \else
\def\thefootnote{}
\footnotetext{%
\textsuperscript{*} Equal Contribution
\hspace{4 mm}
\textsuperscript{\Letter} Corresponding Author}
\fi

\begin{figure}
  \ifreview \vspace{-8 mm} \else \vspace{-10 mm} \fi
  \includegraphics[width=\linewidth]{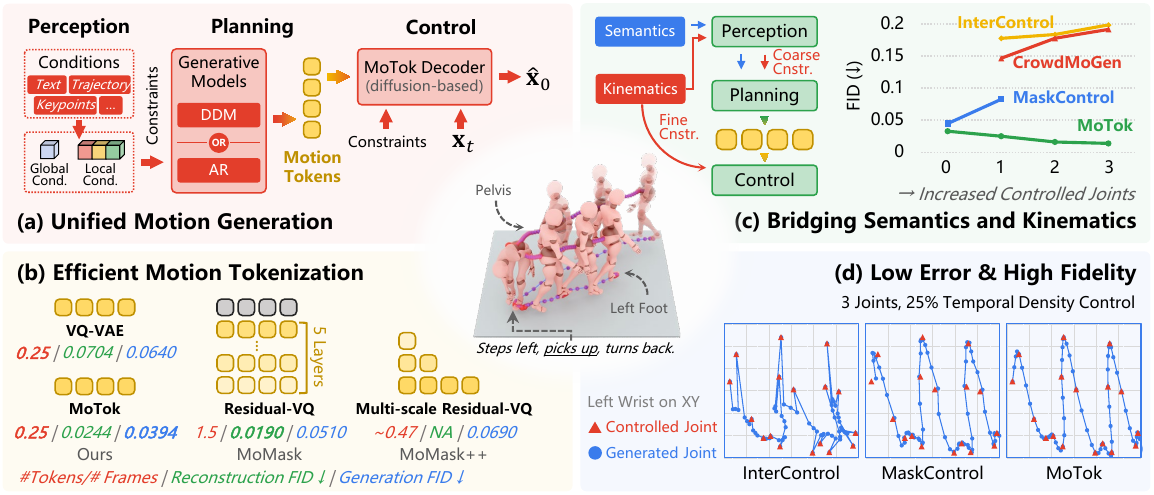}
  \ifreview \else \vspace{-6 mm} \fi
  \caption{%
  \textbf{(a)} A unified Perception–Planning–Control pipeline for conditional motion generation.
  \textbf{(b)} \OM enables compact motion tokenization with fewer tokens while maintaining competitive performance.  
  \textbf{(c)} Bridging semantics and kinematics by applying coarse constraints in planning and fine constraints in motion control. \OM maintains and improves fidelity as more joints are controlled, rather than compromising between controllability and realism.
  \textbf{(d)} Left-wrist XY trajectories with sparse control (red triangles). \OM yields the most natural trajectory and best alignment.}
  \label{fig:teaser}
  \ifreview \vspace{-8 mm} \else \vspace{-12 mm} \fi
\end{figure}

\input{sections/00_abstract}

\input{sections/01_introduction}
\input{sections/02_related_work}
\input{sections/03_method}
\input{sections/04_experiments}
\input{sections/05_conclusion}

\ifreview \else
\input{sections/06_acknowledgement}
\fi

\clearpage
\bibliography{references}
\bibliographystyle{splncs04}

\clearpage
\appendix
\onecolumn
\input{sections/99_appendix}

\end{document}

%% file: sections/00_abstract.tex
\begin{abstract}
Prior motion generation largely follows two paradigms: continuous diffusion models that excel at kinematic control, and discrete token-based generators that are effective for semantic conditioning. 
To combine their strengths, we propose a three-stage framework comprising condition feature extraction \emph{(Perception)}, discrete token generation \emph{(Planning)}, and diffusion-based motion synthesis \emph{(Control)}.
Central to this framework is \textbf{\OM}, a diffusion-based discrete motion tokenizer that decouples semantic abstraction from fine-grained reconstruction by delegating motion recovery to a diffusion decoder, enabling compact single-layer tokens while preserving motion fidelity.
For kinematic conditions, coarse constraints guide token generation during \emph{planning}, while fine-grained constraints are enforced during \emph{control} through diffusion-based optimization.
This design prevents kinematic details from disrupting semantic token planning.
On HumanML3D, our method significantly improves controllability and fidelity over MaskControl while using only \emph{one-sixth of the tokens}, reducing trajectory error from 0.72 cm to 0.08 cm and FID from 0.083 to 0.029.
Unlike prior methods that degrade under stronger kinematic constraints, ours improves fidelity, reducing FID from 0.033 to 0.014.

\keywords{Motion Tokenization \and Diffusion Models \and Conditional Motion Generation}

\end{abstract}

%% file: sections/01_introduction.tex
\section{Introduction}

Human motion generation underpins applications ranging from animation to robotics and embodied agents~\cite{DBLP:preprint/arxiv/2505-05474,DBLP:journals/corr/abs-2601-22153}. 
While recent conditional generative models~\cite{DBLP:conf/iclr/TevetRGSCB23,DBLP:conf/cvpr/GuoMJW024} enable realistic synthesis from high-level semantic inputs, practical scenarios often require additional fine-grained, time-varying kinematic control signals. 
Effectively integrating such low-level constraints while preserving semantic intent remains a central challenge.

Token-based motion generation~\cite{DBLP:conf/cvpr/ZhangZCZZLSY23,DBLP:conf/cvpr/GuoMJW024} compresses continuous motion into discrete tokens for conditional sequence modeling, enabling scalable architectures, flexible conditioning, and the reuse of language-model-style generators. 
However, existing motion tokenizers~\cite{DBLP:conf/cvpr/GuoMJW024} often entangle high-level semantics with low-level motion details, requiring high token rates or hierarchical codes to ensure faithful reconstruction. 
This increases the burden on downstream generators and complicates controllable generation, as fine-grained kinematic condition signals may compete with or override semantic conditioning. 
In contrast, diffusion models~\cite{DBLP:journals/pami/ZhangCPHGYL24,DBLP:conf/cvpr/ChenJLHFCY23} excel at reconstructing continuous motion with smooth dynamics and rich local details. 
This suggests a division of labor in motion generation, where diffusion handles fine-grained reconstruction while discrete tokens capture semantic abstraction.

Motivated by this insight, we propose a \textbf{Perception--Planning--Control} paradigm for controllable motion generation (Fig.~\ref{fig:teaser}\hyperref[fig:teaser]{a}).
In \textbf{Perception}, heterogeneous conditions are encoded as either global conditions (\eg, text) that guide the overall motion, or local conditions (\eg, keypoint trajectories) that provide local constraints.
In \textbf{Planning}, a token-space planner predicts a discrete motion token sequence under a unified interface supporting both \emph{autoregressive} (AR) and \emph{discrete diffusion} (DDM) generators.
In \textbf{Control}, the final motion is synthesized via diffusion-based decoding while enforcing fine-grained kinematic constraints during denoising.
This decomposition separates high-level planning from low-level kinematics, enabling the same pipeline to generalize across generator architectures and motion generation tasks.


Building on this paradigm, we introduce \textbf{\OM}, a diffusion-based discrete motion tokenizer that decouples semantic abstraction from low-level reconstruction.
\OM employs a single-layer codebook to produce compact token sequences (Fig.~\ref{fig:teaser}\hyperref[fig:teaser]{b}), while delegating motion recovery to a diffusion decoder.
This design reduces the token budget for downstream planners and enables decoding-time refinement without forcing discrete tokens to encode fine-grained kinematic details.
Furthermore, we propose a condition injection scheme that harmonizes semantic cues and kinematic constraints by distributing control across stages (Fig.~\ref{fig:teaser}\hyperref[fig:teaser]{c}). 
Kinematic conditions act as \emph{coarse} constraints during the \emph{Planning} stage to guide token generation, and as \emph{fine-grained} constraints during the \emph{Control} stage via optimization-based guidance in diffusion denoising. 
This coarse-to-fine design prevents low-level kinematic details from interfering with token-space planning, avoiding a compromise between controllability and realism.



We evaluate our framework on text-and-trajectory controllable motion generation on HumanML3D~\cite{DBLP:conf/cvpr/GuoZZ0JL022}. 
Compared with MaskControl~\cite{DBLP:conf/iccv/Pinyoanuntapong25}, our method substantially improves both controllability and fidelity, reducing trajectory error from $0.72$ cm to $0.08$ cm and FID from $0.083$ to $0.029$ while using only one-sixth of the tokens.
As shown in Fig.~\ref{fig:teaser}, prior methods~\cite{DBLP:conf/nips/000100L024, DBLP:conf/iccv/Pinyoanuntapong25, DBLP:journals/ijcv/CaoGZXGL26} degrade as more joints are controlled, whereas ours improves motion fidelity under stronger constraints.
Beyond controllable generation, \OM also improves standard text-to-motion performance on HumanML3D under aggressive compression, achieving lower FID than strong token-based baselines with substantially fewer tokens (\eg, $0.039$ vs.\ $0.045$ with one-sixth tokens).


The contributions are summarized as follows:
\vspace{-1 mm}
\begin{enumerate}
\item We propose a three-stage Perception--Planning--Control paradigm for controllable motion generation that supports both autoregressive (AR) and discrete diffusion (DDM) planners under a unified interface.

\item We introduce \OM, a diffusion-based discrete motion tokenizer that decouples semantic abstraction from low-level reconstruction by delegating motion recovery to diffusion decoding, enabling compact single-layer tokens with a dramatically reduced token budget.

\item We develop a coarse-to-fine conditioning scheme that injects kinematic signals as coarse constraints during token planning and enforces fine-grained constraints during diffusion denoising, improving controllability and fidelity.
\end{enumerate}




%% file: sections/02_related_work.tex
\section{Related Work}

\ifreview
\paragraph{Motion Generative Model}
\else
\noindent \textbf{Motion Generative Model.}
\fi
Early motion generation research primarily focused on unconditional settings, with classical methods such as PCA~\cite{DBLP:journals/ivc/OrmoneitBHK05} and Motion Graphs~\cite{DBLP:journals/tog/MinC12}, followed by learning-based generative models including VAEs~\cite{DBLP:conf/mm/GuoZWZSDG020,DBLP:conf/iccv/PetrovichBV21}, implicit functions~\cite{DBLP:conf/eccv/CervantesSSS22}, GANs~\cite{DBLP:conf/cvpr/BarsoumKL18,DBLP:journals/tog/HarveyYNP20}, and normalizing flows~\cite{DBLP:journals/tog/HenterAB20}.
Subsequent text- and action-conditioned approaches~\cite{DBLP:conf/3dim/AhujaM19,DBLP:conf/eccv/PetrovichBV22,DBLP:conf/cvpr/GuoZZ0JL022,DBLP:conf/eccv/TevetGHBC22,DBLP:conf/iccv/PetrovichBV23} aligned motion and language representations via latent-space objectives, but often suffered from limited motion fidelity.
Diffusion-based methods~\cite{DBLP:conf/iclr/TevetRGSCB23,DBLP:journals/pami/ZhangCPHGYL24,DBLP:conf/iccv/ZhangGPCHLYL23,DBLP:journals/corr/abs-2510-26794} significantly improved generation quality through iterative denoising~\cite{DBLP:conf/nips/HoJA20}, yet incur slow inference due to operating on raw motion sequences, 
while latent diffusion~\cite{DBLP:conf/cvpr/RombachBLEO22,DBLP:conf/cvpr/ChenJLHFCY23} accelerates generation at the cost of fine-grained details and editability.
Autoregressive token-based models~\cite{DBLP:conf/cvpr/ZhangZCZZLSY23,DBLP:conf/iccv/ZhongHZX23,DBLP:conf/nips/JiangCLYYC23} further enhance controllability but introduce high computational overhead and limited bidirectional dependency modeling.
Motivated by recent advances in masked modeling~\cite{DBLP:conf/cvpr/GuoMJW024,DBLP:conf/cvpr/Pinyoanuntapong24,DBLP:conf/iccv/Pinyoanuntapong25}, recent works explore efficient and editable motion generation through discrete representations. MaskControl~\cite{DBLP:conf/iccv/Pinyoanuntapong25} designs a differentaible sampling strategy for discrete motion diffusion model, which can achieve spatio-temporal low-level control.
%


\paragraph{Motion Tokenizer}
Early discrete text-to-motion methods such as TM2T~\cite{DBLP:conf/eccv/GuoZWC22} introduce motion tokens by framing motion as a foreign language and learning text–motion translation with VQ-based tokenizers. 
Subsequent works advance along two main directions. 
One line focuses on tokenizer and generator design, improving convolutional tokenizers (\eg, T2M-GPT~\cite{DBLP:conf/cvpr/ZhangZCZZLSY23}), modeling full-body structure more explicitly (\eg, HumanTOMATO~\cite{DBLP:conf/icml/LuCZLZ0S24}), or extending tokenization to the spatio-temporal domain (\eg, MoGenTS~\cite{DBLP:conf/nips/0001HSD0DBH24}), often at the cost of increased modeling complexity.
The other line explores improved quantization schemes. 
MoMask~\cite{DBLP:conf/cvpr/GuoMJW024} introduces residual vector quantization to reduce reconstruction error but substantially increases token count and requires specialized generators, while later variants such as ScaMo~\cite{DBLP:conf/cvpr/Lu0LCDDD0Z25} and MoMask++~\cite{DBLP:conf/nips/GuoHWZ25} investigate alternative or hierarchical quantization strategies to balance efficiency and accuracy.
Despite these advances, existing approaches still face a fundamental trade-off between token efficiency and generation quality, and remain limited in supporting fine-grained, low-level control.

%% file: sections/03_method.tex
\begin{figure*}[t]
  \includegraphics[width=\textwidth]{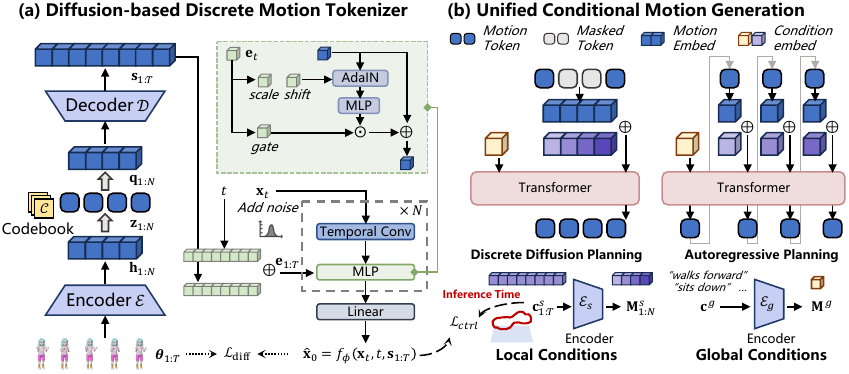}
  \caption{\textbf{Overview of \OM and the unified motion generation framework.} \textbf{(a)} \OM factorizes motion representation into compact discrete tokens and diffusion-based reconstruction by decoding tokens into per-frame conditioning for conditional diffusion. \textbf{(b)} A unified conditional generation framework built on \OM supports both discrete diffusion and autoregressive planners, integrating global and local conditions in a generator-agnostic manner.}
  \label{fig:arch}
  \vspace{-5 mm} 
\end{figure*}

\section{Our Approach}

\subsection{Problem Formulation}

\paragraph{Motion Representation}
A motion sequence is denoted as $\boldsymbol{\theta}_{1:T} = \{\boldsymbol{\theta}_t\}_{t=1}^{T}$, 
where $T$ is the sequence length and $\boldsymbol{\theta}_t \in \mathbb{R}^{D}$ represents the motion state at time $t$. 
The motion state can be instantiated using standard skeleton-based representations commonly adopted in text-to-motion benchmarks (\eg, joint rotations or positions with auxiliary signals), while our framework remains agnostic to the specific choice of parameterization.

\paragraph{Discrete Token Sequence}
Each motion sequence is encoded as a shorter discrete token sequence $\mathbf{\mathbf{z}}_{1:N} = \{\mathbf{z}_n\}_{n=1}^{N}$, where each token $z_n \in {1, \dots, K}$ indexes a shared codebook of size $K$. 
The token compression ratio is defined as $\rho = T / N$. 
A central goal of this work is to achieve high-quality motion generation under aggressive compression (large $\rho$), thereby reducing the sequence modeling burden of downstream generators.

\paragraph{Conditions and Taxonomy}
Heterogeneous conditioning signals $\mathbf{c}$ are categorized into two types:
\textbf{1) Global conditions} $\mathbf{c}^{g}$ provide sequence-level guidance without requiring frame-wise alignment, such as text descriptions or style labels;
and
\textbf{2) local conditions} $\mathbf{c}^{s}_{1:T} = \{\mathbf{c}^{s}_t\}_{t=1}^{T}$ are aligned with the motion timeline and specify kinematic control signals, including target root trajectories, keyframes, contact hints, or motion rhythm.
This taxonomy is used throughout the method to integrate semantic guidance and kinematic constraints in a unified and generator-agnostic manner.

\paragraph{Task Instantiations}
Our formulation supports a range of conditional motion generation tasks, with additional task definitions and experimental results provided in the supplementary material. 
In the main text, we consider two representative settings:
\textbf{1) text-to-motion}, where text provides the global condition $\mathbf{c}^{g}$ and no local conditions $\mathbf{c}^{s}_{1:T}$ is given
and 
\textbf{2) text-and-trajectory control}, where text serves as $\mathbf{c}^{g}$ and a target trajectory is specified as the local conditions $\mathbf{c}^{s}_{1:T}$, requiring the generated motion to follow the trajectory while maintaining semantic consistency.


\subsection{Diffusion-based Discrete Motion Tokenizer}

\OM is a diffusion-based discrete motion tokenizer that factorizes motion representation into a compact discrete code sequence and a diffusion decoder for fine-grained reconstruction. 
%
Unlike conventional VQ-VAE tokenizers that directly decode continuous motion from discrete codes, \OM first maps the discrete codes to a per-frame conditioning signal and then employs a conditional diffusion model to reconstruct motion details. 
By explicitly offloading fine-grained reconstruction to diffusion-based decoding, discrete tokens are freed to focus on semantic structure, enabling a substantially reduced token budget.

As shown Fig.~\ref{fig:arch}\hyperref[fig:arch]{a}, \OM consists of three components:
\textbf{1)} a convolutional encoder $\mathcal{E}(\cdot)$ that produces a temporally downsampled latent sequence;
\textbf{2)} a vector quantizer $\mathcal{Q}(\cdot)$ that maps latents to discrete codes;
and 
\textbf{(3)} a decoder with diffusion-based reconstruction, comprising a convolutional decoder $\mathcal{D}(\cdot)$ and a conditional diffusion model $\mathcal{P}_\phi(\cdot)$.

\paragraph{Convolutional Encoder}
A convolutional encoder $\mathcal{E}(\cdot)$ is used to obtain a compressed latent representation. 
Given a motion sequence $\boldsymbol{\theta}_{1:T}$, latent features are extracted through progressive temporal downsampling:
\begin{equation}
  \mathbf{h}_{1:N} = \mathcal{E}(\boldsymbol{\theta}_{1:T}), \quad \mathbf{h}_{1:N} \in \mathbb{R}^{N \times d},
\end{equation}
where $d$ denotes the latent dimension. 
The temporal length $N$ is determined by the encoder downsampling factor $r$.

\paragraph{Vector Quantizer}
A vector quantization (VQ) module $\mathcal{Q}(\cdot)$ is applied to discretize the latent sequence $\mathbf{h}_{1:N}$.
Let the codebook be $\mathcal{C} = \{\mathbf{c}\}_{k=1}^{K}$, where $K = 1024$ and each code $\mathbf{c}_k \in \mathbb{R}^{d}$.
Each latent vector is assigned to its nearest codebook entry:
\begin{equation}
  \mathbf{z}_n = \arg\min_{k\in\{1,\dots,K\}} \|\mathbf{h}_n - \mathbf{c}_k\|_2^2,
  \quad
  \mathbf{q}_n = \mathbf{c}_{z_n},
\end{equation}
yielding a discrete token sequence $\mathbf{z}_{1:N}$ and the quantized latents $\mathbf{q}_{1:N}$.

\paragraph{Decoder with Diffusion-based Reconstruction}
Rather than directly regressing $\boldsymbol{\theta}_{1:T}$ from the quantized latents $\mathbf{q}_{1:N}$, 
\OM decodes $\mathbf{q}_{1:N}$ into a per-frame conditioning sequence and reconstructs motion using conditional diffusion. 
Specifically, a convolutional decoder $\mathcal{D}(\cdot)$ upsamples the quantized latents as:
\begin{equation}
  \mathbf{s}_{1:T} = \mathcal{D}(\mathbf{q}_{1:N}),
  \quad \mathbf{s}_{1:T}\in\mathbb{R}^{T\times d},
\end{equation}
where $\mathbf{s}_{1:T}$ serves as the conditioning signal for diffusion-based reconstruction.
We then define a conditional diffusion decoder as a reverse diffusion process
$\mathcal{P}_\phi(\cdot)$ parameterized by a neural denoiser $f_\phi$.
Concretely, $f_\phi$ predicts the clean motion $\hat{\mathbf{x}}_0$ from a noisy input $\mathbf{x}_t$ at diffusion timestep $t$:
\begin{equation}
  \hat{\mathbf{x}}_0 = f_{\phi}(\mathbf{x}_t, t, \mathbf{s}_{1:T}).
\end{equation}
This prediction defines the reverse transitions of the diffusion model, yielding a distribution
$p_\phi(\mathbf{x}_{t-1}\mid \mathbf{x}_t, \mathbf{s}_{1:T})$.
At inference, $\mathcal{P}_\phi$ samples the reconstructed motion by iteratively applying reverse steps from
$\mathbf{x}_T\sim\mathcal{N}(\mathbf{0}, \mathbf{I})$ to obtain $\hat{\mathbf{x}}_0$.
Architecturally, $f_{\phi}$ first projects $\mathbf{x}_t$ to the latent dimension via a linear layer, followed by a stack of processing blocks.
Each block contains a residual 1D convolution module for enhanced temporal modeling and an MLP that injects conditioning embeddings $\mathbf{e}_{1:T}$ into motion features via an AdaIN-style transformation, where $\mathbf{e}_{1:T}$ combines timestep embeddings with conditioning signals $\mathbf{s}_{1:T}$.
This diffusion-based decoding provides a natural interface for enforcing additional fine-grained constraints during reconstruction (\eg, trajectories or joint-level hints), as such constraints can be applied throughout the denoising process rather than solely being imposed at the level of discrete token prediction.

\paragraph{Training Objectives}
\OM is trained end-to-end using a combination of a diffusion reconstruction objective and a VQ commitment loss, following the diffusion training strategy of MAR~\cite{DBLP:conf/nips/LiTLDH24}.
During diffusion training, a timestep $t$ is sampled and the conditional denoising objective is optimized:
\begin{equation}
\mathcal{L}_{\mathrm{diff}} = \mathbb{E}_{t,\epsilon}\Big[
\ell\big(\hat{\mathbf{x}}_0, \mathbf{x}_0\big)\Big],
\end{equation}
where $\ell(\cdot,\cdot)$ denotes the Smooth-$\ell_1$ loss and $\mathbf{x}_0 = \boldsymbol{\theta}_{1:T}$ is the clean motion sequence.
In addition, the VQ commitment loss returned by the quantizer is included with weight $\lambda_{\mathrm{commit}} = 0.02$, yielding the overall training objective:
\begin{equation}
  \mathcal{L} = \mathcal{L}_{\mathrm{diff}} + \lambda_{\mathrm{commit}}\mathcal{L}_{\mathrm{commit}}.
\end{equation}

\subsection{Unified Conditional Motion Generation}

As shown in Fig.~\ref{fig:arch},
\OM enables a unified conditional motion generation pipeline by decoupling \textbf{planning in discrete token space} from \textbf{control in diffusion-based decoding}. 
Given a condition set $\mathbf{c}=\{\mathbf{c}^g,\mathbf{c}^s_{1:T}\}$, a token generator first produces a discrete sequence $\mathbf{z}_{1:N}$, which is then decoded into a continuous motion $\hat{\mathbf{x}}_0$ via diffusion conditioned on features derived from \OM. 
This formulation supports both discrete diffusion and autoregressive token generators through a shared conditioning interface.

Conditions are categorized by their temporal characteristics into \textbf{1) global conditions} $\mathbf{c}^g$, which provide sequence-level guidance without frame-wise alignment (\eg, text descriptions), 
and 
\textbf{2) local conditions} $\mathbf{c}^s_{1:T}$, which are aligned with the motion timeline and specify fine-grained control signals (\eg, target trajectories). 
Global conditions are encoded by $\mathcal{E}_g(\cdot)$ into a sequence-level feature $\mathbf{M}^g \in \mathbb{R}^{d}$ and used as a dedicated token during discrete planning, while local conditions are encoded by $\mathcal{E}_s(\cdot)$ into a feature sequence $\mathbf{M}_{1:N}^s \in \mathbb{R}^{N\times d}$ aligned with the token length $N$.

\paragraph{Planning in Discreate Token Space}
Token-space planning generates discrete motion tokens under heterogeneous conditions and supports both discrete diffusion and autoregressive generators through a shared planning interface.

\noindent \textbf{Discrete Diffusion Planning} follows the masked-token diffusion paradigm introduced by MoMask~\cite{DBLP:conf/cvpr/GuoMJW024}, where subsets of tokens are iteratively predicted conditioned on observed tokens and external conditions. 
To inject conditions in a unified manner, a token embedding sequence of length $N+1$ is constructed, with the first position reserved for the global condition feature and the remaining $N$ positions corresponding to motion tokens:
\begin{equation}
  \mathbf{H}_0 = [\,\mathbf{M}^g;\, \mathbf{h}_1;\dots;\mathbf{h}_N\,],
\end{equation}
where $\mathbf{h}_n$ denotes the learnable embedding of token $z_n$ or a learned \texttt{[MASK]} embedding for masked positions. 
local conditions features are incorporated by additive fusion with positional embeddings at the motion-token positions:
\begin{equation}
  \mathbf{H}_0[1+n] \leftarrow \mathbf{H}_0[1+n] + \mathbf{M}_n^s + \mathbf{p}_n,\quad n=1,\dots,N,
\end{equation}
where $\mathbf{p}_n$ is the standard positional embedding. 
$\mathbf{M}^g$ at position $1$ attends to all motion tokens, providing sequence-level guidance throughout denoising.

\noindent \textbf{Autoregressive Planning} follows the same interface, with the global condition occupying the first position and motion tokens generated sequentially in a causal manner, as in T2M-GPT~\cite{DBLP:conf/cvpr/ZhangZCZZLSY23}. 
Due to the one-step shift inherent to next-token prediction, the local conditions embedding for the first token is added to the global-conditioning position, while the embedding for each subsequent token is added to the preceding token position. 
This design preserves temporal alignment of control signals and allows \OM to be integrated into autoregressive backbones with minimal modification.

\noindent \textbf{Classifier-free Guidance} (CFG) is applied to token-space planning and extended to multiple conditions via alternating guidance pairs, following ReMoDiffuse~\cite{DBLP:conf/iccv/ZhangGPCHLYL23}. 
Let $\epsilon_{\theta}(\mathbf{z}; \mathbf{c})$ denote the sampling output of the token generator under conditions $\mathbf{c}={\mathbf{c}^g,\mathbf{c}^s_{1:T}}$. 
For single-condition CFG, conditional and unconditional predictions are formed with $\mathbf{c}{\text{cond}}={\mathbf{g}, \varnothing}$ and $\mathbf{c}{\text{uncond}}={\varnothing, \varnothing}$, and are combined as
\begin{equation}
\epsilon_{\text{cfg}} = \epsilon_{\theta}(\mathbf{z}; \mathbf{c}_{\text{uncond}})
+ w\Big(\epsilon_{\theta}(\mathbf{z}; \mathbf{c}_{\text{cond}}) - \epsilon_{\theta}(\mathbf{z}; \mathbf{c}_{\text{uncond}})\Big),
\end{equation}
where $w$ is the guidance scale.
When both semantic and trajectory conditions are present, fully dropping conditions in the unconditional branch may bias generation toward a single modality. 
To balance semantic guidance and control fidelity, two CFG pairs are alternated with equal probability:
\begin{align}
\text{(A)}\quad & \mathbf{c}_{\text{cond}}=\{\mathbf{g}, \mathbf{s}\}, \qquad
\mathbf{c}_{\text{uncond}}=\{\varnothing, \mathbf{s}\},\\
\text{(B)}\quad & \mathbf{c}_{\text{cond}}=\{\mathbf{g}, \mathbf{s}\}, \qquad
\mathbf{c}_{\text{uncond}}=\{\mathbf{g}, \varnothing\}.
\end{align}
The same CFG combination rule is applied. 
This alternating strategy enables effective multi-condition guidance during planning without introducing additional networks or training objectives.

\paragraph{Control in Diffusion Decoding}
After token-level planning, discrete tokens are decoded by \OM into a per-frame conditioning sequence $\mathbf{s}_{1:T}$, and motion is reconstructed via conditional diffusion. 
Fine-grained control is enforced directly during denoising by optimizing an auxiliary control objective. 
At diffusion step $k$, given the current estimate $\hat{\mathbf{x}}_k$ of the full motion sequence, a control loss $\mathcal{L}_{\text{ctrl}}(\hat{\mathbf{x}}_k, \mathbf{c}^s_{1:T})$ measures deviation from local conditions (\eg, trajectory adherence), and the denoising update is refined via
\begin{equation}
  \hat{\mathbf{x}}_{k} \leftarrow
  \hat{\mathbf{x}}_{k} - \eta \nabla_{\hat{\mathbf{x}}_k} \mathcal{L}_{\text{ctrl}}(\hat{\mathbf{x}}_{k}, \mathbf{c}^s_{1:T}),
\end{equation}
where $\eta$ controls the refinement strength. 
Enforcing constraints at the continuous-motion level enables precise low-level control without burdening the discrete planner with high-frequency details, and is critical for achieving both low trajectory error and improved motion fidelity when semantic and low-level conditions are jointly applied.


\subsection{Instantiation for Different Tasks}

The unified framework is instantiated for two representative tasks, with additional tasks and results provided in the Appendix. 
Unless otherwise specified, global conditions are encoded into sequence-level features, while local conditions are encoded into token-aligned features. 
Both discrete diffusion and autoregressive token generators are supported through the same conditioning interface.


\paragraph{Text-to-Motion}
In text-to-motion generation, conditioning is purely global. 
Given a text prompt $\mathbf{t}$, a sequence-level embedding is extracted using a pretrained CLIP text encoder~\cite{DBLP:conf/icml/RadfordKHRGASAM21}:
\begin{equation}
  \mathbf{M}^g = \mathcal{E}_{\text{text}}(\mathbf{t}) \in \mathbb{R}^{d}.
  \label{eq:mg_text}
\end{equation}
%


\paragraph{Text and Trajectory Control}
For joint text-and-trajectory generation, a global text embedding is combined with a time-synchronized trajectory embedding. 
Given a text prompt $\mathbf{t}$ and a target trajectory $\boldsymbol{\tau}_{1:T} \in \mathbb{R}^{T \times J \times 3}$, 
where $J$ denotes the number of joints, 
the text prompt is encoded into a global feature $\mathbf{M}^g$ as in Eq.~\ref{eq:mg_text}, 
while the trajectory is encoded into a token-aligned sequence using the same convolutional encoder as in \OM:
\begin{equation}
  \mathbf{M}_{1:N}^s = \mathcal{E}_\text{traj}(\boldsymbol{\tau}_{1:T}) \in \mathbb{R}^{N\times d}.
\end{equation}
The trajectory features are injected as local conditions during token-space planning and further enforced during diffusion decoding through refinement. This design allows semantic planning to occur in token space, while precise trajectory adherence is handled at the continuous-motion level.


%% file: sections/04_experiments.tex
\section{Experiments}
\input{tables/quant_traj}

\subsection{Experimental Setup}

\paragraph{Datasets}
Experiments are conducted on HumanML3D~\cite{DBLP:conf/cvpr/GuoZZ0JL022} and KIT-ML~\cite{DBLP:journals/bigdata/PlappertMA16}, which are widely used paired text--motion benchmarks.
Each dataset provides natural language descriptions for every motion sequence, together with a standardized skeleton-based motion representation.


\paragraph{Text Conditioning}
Text conditions are treated as global conditions and encoded into a sequence-level feature using a pretrained CLIP text encoder (ViT-B/32)~\cite{DBLP:conf/icml/RadfordKHRGASAM21}, producing a 512-dimensional embedding. 
This embedding serves as the global condition token during token-space planning for both discrete diffusion and autoregressive generators.


\paragraph{Trajectory Conditioning}
Trajectory conditions are treated as local conditions and encoded into token-aligned feature sequences using a convolutional encoder that mirrors the motion encoder in \OM. 
The encoder downsamples the trajectory to match the token length, with dropout (0.1) applied for robustness and classifier-free guidance.


\paragraph{Training}
\OM is evaluated with two token-space planning backbones: 
\textbf{1)} a MoMask-style~\cite{DBLP:conf/cvpr/GuoMJW024} discrete diffusion planner (dim $384$, $6$ layers) and 
\textbf{2)} a T2M-GPT-style~\cite{DBLP:conf/cvpr/ZhangZCZZLSY23} autoregressive transformer (dim $768$, $9$ layers).
During training, condition dropout is applied with probability $0.1$, and random token replacement with probability $0.1$ (DDM) / $0.2$ (AR).
Four variants are reported: \OM-DDM-4/2 and \OM-AR-4/2, where DDM/AR denote the generative model and 4/2 the temporal compression ratio.



\paragraph{Inference}
During reconstruction and controllable generation, discrete tokens are decoded by \OM and converted into continuous motion using conditional diffusion. 
When trajectory control is enabled, decoding-time refinement is applied by incorporating an auxiliary control objective at each denoising step, following the strategy of InterControl~\cite{DBLP:conf/nips/000100L024}.

\begin{figure}[!t]
  \includegraphics[width=\linewidth]{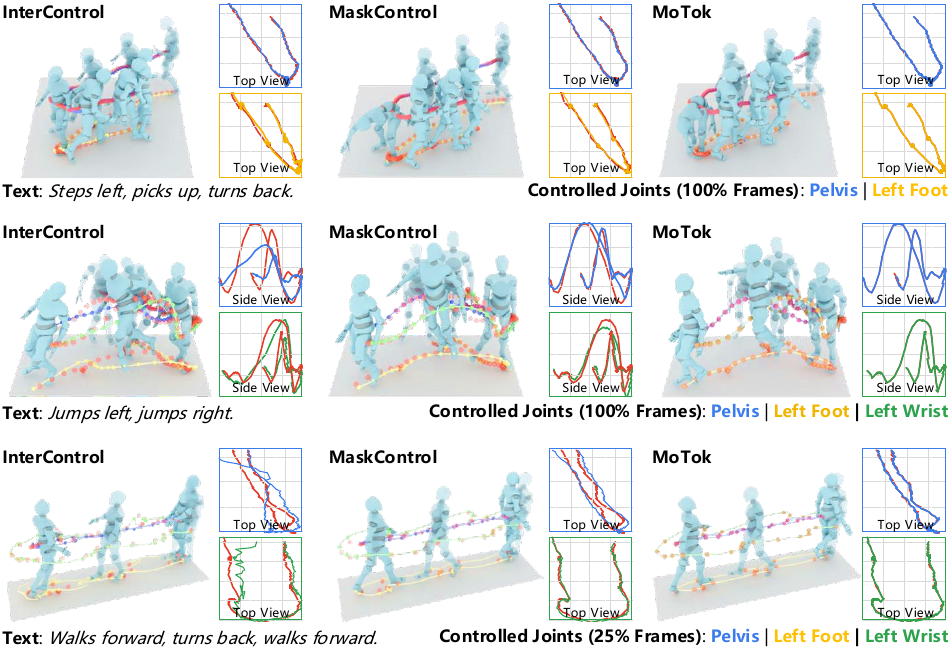}
  \caption{\textbf{Visual comparison with state-of-the-art methods for any-joint any-frame control.} The right panels show trajectory views of two of the controlled joints. \textcolor{red}{Red} indicates the control signal and \textcolor{blue}{blue} indicates the generated motion.}
  \label{fig:motion-control}
  \vspace{-6 mm}
\end{figure}

\subsection{Text and Trajectory Control}

Table~\ref{tab:quant_traj} shows that \OM achieves substantially improved controllable generation on HumanML3D. 
Compared to the strongest DDM-based baseline, \OM-DDM-4 reduces FID from 0.061 to 0.027 while nearly halving trajectory error (0.098 to 0.049) under the ``Pelvis'' setting. 
The improvement is even larger in the Random One'' setting, where \OM-DDM-2 reduces FID from 0.083 to 0.025 and average error from 0.0072 to 0.0008.
Notably, these gains are achieved with substantially fewer tokens: \OM uses only 1/6 of MaskControl's token budget in the first setting and 1/3 in the second, demonstrating its superior efficiency.

While prior methods such as MaskControl~\cite{DBLP:conf/iccv/Pinyoanuntapong25} degrade in motion quality when trajectory constraints are introduced (FID 0.061 vs. 0.045 from MoMask in Table~\ref{tab:quant_t2m}), both \OM-DDM and \OM-AR achieve lower FID under joint text-and-trajectory control than in text-only generation.
This suggests that trajectory information acts as complementary guidance rather than a competing constraint.
This improvement stems from
\textbf{1)} a condition hierarchy that separates global semantic constraints from fine-grained trajectory constraints during token-space planning, and
\textbf{2)} decoding-time diffusion guidance that enforces low-level constraints without overloading the discrete planner.

Figure~\ref{fig:motion-control} presents a qualitative comparison across several methods. As shown, MoTok follows the provided trajectory most faithfully while producing smooth overall motion. Although MaskControl also yields high-quality motions, its trajectory alignment is limited by the capacity of its tokenizer, making it difficult to match fine-grained trajectory details.

Overall, \OM better aligns semantic planning with low-level control, producing motions that are both more trajectory-accurate and closer to the real motion distribution.
Although the AR variants perform slightly worse than the DDM versions, likely due to weaker global planning capacity, they still outperform prior methods, demonstrating the generality of the \OM paradigm.

\subsection{Text-to-Motion Generation}

As shown in Table~\ref{tab:quant_t2m}, we evaluate \OM on two standard text-to-motion benchmarks, HumanML3D and KIT-ML, where it achieves strong overall performance, particularly in motion realism measured by FID.

On HumanML3D, \OM-DDM consistently outperforms the MoMask baseline under significantly smaller token budgets: with only \textbf{one-sixth of the tokens}, \OM-DDM-4 attains a lower FID than MoMask (0.039 vs. 0.045), and with higher capacity, \OM-DDM-2 further reduces FID to 0.033, the best among all compared methods.
\OM also transfers effectively to autoregressive planning. Under the same token budget, \OM-AR-4 reduces the FID of T2M-GPT by \textbf{nearly threefold} (0.053 vs. 0.141), indicating a substantially reduced modeling burden for token-space generators.

On KIT-ML, \OM achieves the lowest FID of 0.144, improving over the strongest baseline (0.155) while maintaining competitive performance on other metrics.
Overall, these results show that \OM produces high-quality motions with substantially fewer tokens, supporting the effectiveness of diffusion-based decoding for compact and semantically expressive motion representations.

\input{tables/quant_t2m}




\subsection{Ablation Study}

Table \ref{tab:quant_ablation} presents an ablation study of tokenizer configurations under both discrete diffusion (DDM) and autoregressive (AR) planners, analyzing decoder design, latent dimension, and temporal downsampling rate.
Beyond raw performance differences, the results reveal a consistent pattern: tokenizer effectiveness is largely determined by how temporal structure is handled under noisy generation, rather than by reconstruction fidelity alone.


\input{tables/ablation_study}

\paragraph{Decoder Design}
%
Across all settings, decoder performance follows a clear hierarchy: diffusion-based decoders outperform plain convolutional decoding, and introducing explicit temporal modeling further improves over frame-wise diffusion.
With the same compression ratio (downrate = 4), a convolutional decoder yields poor reconstruction (Recon FID = 0.0704), whereas diffusion-based decoding reduces Recon FID to below 0.04, demonstrating its advantage in recovering fine-grained motion details.
A lightweight DiffusionHead \cite{DBLP:conf/nips/LiTLDH24} achieves strong reconstruction and is sufficient for controllable generation, but offers limited benefit for the planning stage. 
This gap reflects the difference between reconstruction on clean tokens and generation under stochastic noise.
In contrast, DiffusionConv introduces local temporal correlations via residual 1D convolutions, consistently improving generation fidelity and controllability under both planners. 

\paragraph{Latent Dimension}
Fixing the decoder and temporal compression (downrate = 4), we vary the codebook dimension in Table~\ref{tab:quant_ablation} (rows 3–8).
A latent dimension of $d=768$ provides sufficient capacity for accurate reconstruction and stable token-space planning, whereas smaller dimensions reduce expressiveness and degrade both generation fidelity and controllability.

\paragraph{Temporal Downsampling Rate}
Varying the temporal downsampling rate under a fixed decoder and latent dimension (Table~\ref{tab:quant_ablation}, rows 9–16) reveals a non-monotonic trade-off.
Low downrates produce dense tokens that are harder to model under noisy generation, while excessive compression removes essential structure.
Moderate downsampling (downrate 2 or 4) consistently yields the best balance between reconstruction quality and downstream planning.

\paragraph{Kernel Size in Temporal Diffusion}
Fixing the latent dimension and downsampling rate (Table~\ref{tab:quant_ablation}, rows 3–6), we observe that kernel size has a non-trivial effect on performance.
A moderate temporal receptive field ($k=5$) yields the best overall results, whereas larger kernels over-smooth sparse tokens and degrade generation quality. We also find that the optimal kernel size is closely tied to the temporal compression ratio. When the compression ratio is low (DR = 1 or 2), where the number of tokens is relatively large, a kernel size of 7 performs better than 5. However, under higher compression (fewer tokens), a kernel size of 5 yields better results. This suggests that a moderate kernel size strikes a better balance between modeling temporal dependencies and avoiding over-smoothing.

\input{tables/ablation_low_ctrl}

\paragraph{Dual-path Low-level Conditioning}
As shown in Table~\ref{tab:control_injection_ablation}, low-level conditioning is necessary and cannot be removed without harming controllable generation. 
Injecting low-level signals only in the generator forces token-space planning to simultaneously satisfy semantics and precise constraints, often increasing control error and degrading motion fidelity. Injecting them only in the tokenizer decoder lacks planning-time guidance and leads to unstable or weaker adherence. The best results are achieved only when low-level conditions are applied in both stages, where the generator provides coarse, condition-aware planning and the diffusion decoder enforces fine-grained constraints during denoising. 

%% file: tables/quant_traj.tex
\begin{table*}[!t]
  \setlength\tabcolsep{4pt}
  \setlength\extrarowheight{1pt}
  \caption{\textbf{Controllable motion generation quantitative results on HumanML3D test set.} 
  Note that ``Traj. Err.'', ``Loc. Err.'', and ``Avg. Err.'' denote ``Trajectory Error'', ``Localization Error'', and ``Average Error'', respectively.
  \uarr(\darr) indicates that the values are better if the metric is larger (smaller). `$\rightarrow$' means closer to real data is better.
  \textcolor{red}{Red} and \textcolor{blue}{Blue} highlight the best and second-best results, respectively.}
  \vspace{-2 mm}
  \label{tab:quant_traj}
  \resizebox{\textwidth}{!}{%
    \begin{tabular}{clccccccc}
    \toprule
        & Method
        & FID \darr
        & \makecell{R-Precision \uarr\\(Top-3)}
        & Diversity \rarr
        & \makecell{Foot Skating\\Ratio \darr}
        & \makecell{Traj. Err.\darr\\(50cm)}
        & \makecell{Loc. Err. \darr\\(50cm)}
        & \makecell{Avg. Err. \darr\\(m)} \\ 
    \midrule
    & Real Motion 
    & 0.002           & 0.797           & 9.503           & 0.000
    & \mean[1]{0.0000}         & \mean[1]{0.0000}         & \mean[1]{0.0000}\\ 
    \midrule
    \multirow{10}{*}{\rotatebox{90}{Pelvis}}
    & PriorMDM~\cite{DBLP:conf/iclr/ShafirTKB24} 
    & 0.475            & 0.583            & 9.156           & 0.0897
    & 0.3457           & 0.2132           & 0.4417 \\
    & GMD~\cite{DBLP:conf/iccv/KarunratanakulP23} 
    & 0.576            & 0.665            & 9.206           & 0.109
    & 0.0931           & 0.0321           & 0.1439  \\
    & OmniControl~\cite{DBLP:conf/iclr/XieJZSJ24} 
    & 0.218            & 0.687            & 9.422           & 0.0547
    & 0.0387           & 0.0096           & 0.0338  \\
    & InterControl~\cite{DBLP:conf/nips/000100L024} 
    & 0.159            & 0.671            & 9.482           & 0.0729
    & \mean[2]{0.0132} & \mean[2]{0.0004} & 0.0496 \\
    & CrowdMoGen~\cite{DBLP:journals/ijcv/CaoGZXGL26} 
    & 0.132            & 0.784            & 9.109           & 0.0762 
    & \mean[1]{0.0000} & \mean[1]{0.0000} & 0.0196 \\
    & MaskControl~\cite{DBLP:conf/iccv/Pinyoanuntapong25} 
    & 0.061            & \mean[1]{0.809}  & \mean[1]{9.496} & 0.0547
    & \mean[1]{0.0000} & \mean[1]{0.0000} & 0.0098 \\
    \cline{2-9}
    & \bf{\OM-AR-2} 
    & 0.046            & 0.767            & \mean[2]{9.516} & \mean[2]{0.0489}
    & \mean[1]{0.0000} & \mean[1]{0.0000} & 0.0052 \\
    & \bf{\OM-DDM-2} 
    & \mean[2]{0.035}  & 0.786            & 9.440           & \mean[1]{0.0453}
    & \mean[1]{0.0000} & \mean[1]{0.0000} & \mean[1]{0.0049} \\
    & \bf{\OM-AR-4} 
    & 0.046            & 0.777            & 9.408           & 0.0605
    & \mean[1]{0.0000} & \mean[1]{0.0000} & \mean[2]{0.0051} \\
    & \bf{\OM-DDM-4} 
    & \mean[1]{0.029}  & \mean[2]{0.794}  & 9.476           & 0.0548
    & \mean[1]{0.0000} & \mean[1]{0.0000} & \mean[1]{0.0049} \\
    \midrule
    \multirow{7}{*}{\rotatebox{90}{Random One}} 
    & OmniControl~\cite{DBLP:conf/iclr/XieJZSJ24} 
    & 0.310            & 0.693            & \mean[1]{9.502} & 0.0608
    & 0.0617           & 0.0107           & 0.0404 \\
    & InterControl~\cite{DBLP:conf/nips/000100L024}
    & 0.178            & 0.669            & \mean[2]{9.498} & 0.0968
    & \mean[2]{0.0403} & \mean[2]{0.0031} & 0.0741 \\
    & CrowdMoGen~\cite{DBLP:journals/ijcv/CaoGZXGL26} 
    & 0.147            & \mean[2]{0.781}  & 9.461           & 0.0829
    & \mean[1]{0.0000} & \mean[1]{0.0000} & 0.0028 \\
    & MaskControl~\cite{DBLP:conf/iccv/Pinyoanuntapong25} 
    & 0.083            & \mean[1]{0.805}  & 9.395           & 0.0545 
    & \mean[1]{0.0000} & \mean[1]{0.0000} & 0.0072 \\
    \cline{2-9}
    & \bf{\OM-AR-2} 
    & 0.056            & 0.744            & 9.586           & \mean[1]{0.0430}
    & \mean[1]{0.0000} & \mean[1]{0.0000} & \mean[2]{0.0009} \\
    & \bf{\OM-DDM-2} 
    & \mean[1]{0.025}  & 0.779            & 9.756           & \mean[2]{0.0436}
    & \mean[1]{0.0000} & \mean[1]{0.0000} & \mean[1]{0.0008} \\
    & \bf{\OM-AR-4} 
    & 0.040            & 0.756            & 9.380           & 0.0496
    & \mean[1]{0.0000} & \mean[1]{0.0000} & \mean[2]{0.0009} \\
    & \bf{\OM-DDM-4} 
    & \mean[2]{0.029}  & 0.777            & 9.297           & 0.0508
    & \mean[1]{0.0000} & \mean[1]{0.0000} & \mean[1]{0.0008} \\
    \midrule
    \multirow{6}{*}{\rotatebox{90}{Random Two}} 
    & InterControl~\cite{DBLP:conf/nips/000100L024} 
    & 0.184            & 0.670            & 9.410           & 0.0948 
    & \mean[2]{0.0475} & \mean[2]{0.0030} & 0.0911 \\
    & CrowdMoGen~\cite{DBLP:journals/ijcv/CaoGZXGL26} 
    & 0.178            & \mean[2]{0.777}  & 9.149           & 0.0865 
    & \mean[1]{0.0000} & \mean[1]{0.0000} & \mean[2]{0.0027} \\  
    \cline{2-9}
    & \bf{\OM-AR-2} 
    & 0.037            & 0.765            & 9.452           & \mean[1]{0.0422}
    & \mean[1]{0.0000} & \mean[1]{0.0000} & \mean[1]{0.0006} \\
    & \bf{\OM-DDM-2} 
    & \mean[1]{0.016}  & 0.775            & 9.711           & \mean[2]{0.0456}
    & \mean[1]{0.0000} & \mean[1]{0.0000} & \mean[1]{0.0006} \\ 
    & \bf{\OM-AR-4} 
    & 0.041            & 0.762            & \mean[1]{9.481} & 0.0515
    & \mean[1]{0.0000} & \mean[1]{0.0000} & \mean[1]{0.0006} \\
    & \bf{\OM-DDM-4} 
    & \mean[2]{0.022}  & \mean[1]{0.784}  & \mean[2]{9.562} & 0.0521
    & \mean[1]{0.0000} & \mean[1]{0.0000} & \mean[1]{0.0006} \\
    \midrule
    \multirow{6}{*}{\rotatebox{90}{Random Three}} 
    & InterControl~\cite{DBLP:conf/nips/000100L024} 
    & 0.199            & 0.673            & 9.352           & 0.0930 
    & \mean[2]{0.0487} & \mean[2]{0.0026} & 0.0969 \\
    & CrowdMoGen~\cite{DBLP:journals/ijcv/CaoGZXGL26} 
    & 0.192            & 0.778            & 9.169           & 0.0871 
    & \mean[1]{0.0000} & \mean[1]{0.0000} & 0.0030 \\    
    \cline{2-9}
    & \bf{\OM-AR-2} 
    & 0.035            & 0.772            & 9.278           & \mean[2]{0.0482}
    & \mean[1]{0.0000} & \mean[1]{0.0000} & \mean[2]{0.0008} \\
    & \bf{\OM-DDM-2} 
    & \mean[1]{0.014}  & \mean[2]{0.786}  & \mean[1]{9.519} & \mean[1]{0.0480}
    & \mean[1]{0.0000} & \mean[1]{0.0000} & \mean[1]{0.0007} \\ 
    & \bf{\OM-AR-4} 
    & 0.045            & 0.769            & \mean[1]{9.519} & 0.0529
    & \mean[1]{0.0000} & \mean[1]{0.0000} & \mean[2]{0.0008} \\
    & \bf{\OM-DDM-4} 
    & \mean[2]{0.022}  & \mean[1]{0.788}  & \mean[2]{9.474} & 0.0523
    & \mean[1]{0.0000} & \mean[1]{0.0000} & \mean[1]{0.0007} \\
    \bottomrule
    \end{tabular}
  }
  \vspace{-4 mm}
\end{table*}

%% file: tables/quant_t2m.tex
\begin{table*}[!t]
  \setlength\tabcolsep{4pt}
  \setlength\extrarowheight{1pt}
  \caption{\textbf{Quantitative results on the HumanML3D and KIT-ML test sets.} For each metric, we conduct the evaluation 20 times and report the average with a 95\% confidence interval. The right arrow (\rarr) signifies that results closer to real motion are better. \textcolor{red}{Red} and \textcolor{blue}{Blue} highlight the best and second-best results, respectively. Notably, \OM-DDM operates with \textbf{a token budget that is only one-sixth} of that used by MoMask, while maintaining comparable or improved performance across most metrics.}
  \vspace{-2 mm}
  \label{tab:quant_t2m}
  \resizebox{\linewidth}{!}{
  \begin{tabular}{llccccccc} 
    \toprule
      \multirow{2}{*}{} & 
      \multirow{2}{*}{Method} & 
      \multicolumn{3}{c}{R-Precision \uarr} & 
      \multirow{2}{*}{FID \darr} & 
      \multirow{2}{*}{MM-Dist \darr} & 
      \multirow{2}{*}{Diversity \rarr} & 
      \multirow{2}{*}{MModality \uarr} \\ 
    \cline{3-5}
      &              &  Top-1 \uarr & Top-2 \uarr & Top-3 \uarr
      &              &              &             & \\ 
    \midrule
    \multirow{9}{*}{\rotatebox{90}{HumanML3D}}
    & Real Motion 
    & \std{0.511}{.003}    & \std{0.703}{.003}    & \std{0.797}{.002}    & \std{0.002}{.000}
    & \std{2.974}{.008}    & \std{9.503}{.065}    & - \\
    \cline{2-9}
    & MotionDiffuse~\cite{DBLP:journals/pami/ZhangCPHGYL24}
    & \std{0.491}{.001}    & \std{0.681}{.001}    & \std{0.782}{.001}    & \std{0.630}{.001}
    & \std{3.113}{.001}    & \std[1]{9.410}{.049} & \std{1.553}{.042}    \\
    & ReMoDiffuse~\cite{DBLP:conf/iccv/ZhangGPCHLYL23}
    & \std{0.510}{.005}    & \std{0.698}{.006}    & \std{0.795}{.004}    & \std{0.103}{.004}
    & \std{2.974}{.016}    & \std{9.018}{.075}    & \std{1.795}{.043}    \\
    & T2M-GPT~\cite{DBLP:conf/cvpr/ZhangZCZZLSY23}
    & \std{0.492}{.003}    & \std{0.679}{.002}    & \std{0.775}{.002}    & \std{0.141}{.005}
    & \std{3.121}{.009}    & \std{9.722}{.082}    & \std{1.831}{.048} \\
    & MoMask~\cite{DBLP:conf/cvpr/GuoMJW024} 
    & \std[1]{0.521}{.002} & \std[1]{0.713}{.002} & \std[1]{0.807}{.002} & \std{0.045}{.002}
    & \std[1]{2.958}{.008} & -                    & \std{1.241}{.040}    \\
    & MotionLCM~\cite{DBLP:conf/eccv/DaiCWLDT24}
    & \std{0.502}{.003}    & \std{0.698}{.002}    & \std{0.798}{.002}    & \std{0.304}{.012}
    & \std{3.012}{.007}    & \std{9.607}{.066}    & \std[2]{2.259}{.092} \\
    \cline{2-9}
    & \bf{\OM-AR-2}  
    & \std{0.483}{.003}    & \std{0.670}{.002}    & \std{0.770}{.002}    & \std{0.061}{.005}
    & \std{3.128}{.011}    & \std{9.675}{0.09}    & \std{2.237}{.050} \\
    & \bf{\OM-DDM-2} 
    & \std[2]{0.512}{.002} & \std[2]{0.705}{.002} & \std[2]{0.799}{.002} & \std[1]{0.033}{.002}
    & \std[2]{2.981}{.010} & \std{9.523}{0.09}    & \std[1]{2.639}{.079} \\
    & \bf{\OM-AR-4}  
    & \std{0.491}{.004}    & \std{0.676}{.002}    & \std{0.773}{.003}    & \std{0.053}{.004}
    & \std{3.106}{.014}    & \std{9.691}{0.07}    & \std{2.183}{.048} \\
    & \bf{\OM-DDM-4}
    & \std{0.500}{.003}    & \std{0.693}{.002}    & \std{0.793}{.002}    & \std[2]{0.039}{.002}
    & \std{3.030}{.007}    & \std[2]{9.411}{.078} & \std[1]{2.639}{.063} \\
    \midrule
    \multirow{8}{*}{\rotatebox{90}{KIT-ML}}
    & Real Motion  
    & \std{0.424}{.005}    & \std{0.649}{.006}    & \std{0.779}{.006}    & \std{0.031}{.004}
    & \std{2.788}{.012}    & \std{11.08}{.097}    & - \\
    \cline{2-9}
    & MotionDiffuse~\cite{DBLP:journals/pami/ZhangCPHGYL24}
    & \std{0.417}{.004}    & \std{0.621}{.004}    & \std{0.739}{.004}    & \std{1.954}{.062}
    & \std{2.958}{.005}    & \std[1]{11.10}{.143} & \std{0.730}{.013}    \\
    & ReMoDiffuse~\cite{DBLP:conf/iccv/ZhangGPCHLYL23}  
    & \std{0.427}{.014}    & \std{0.641}{.004}    & \std[2]{0.765}{.055} & \std[2]{0.155}{.006}
    & \std[2]{2.814}{.012} & \std{10.80}{.105}    & \std{1.239}{.028} \\
    & T2M-GPT~\cite{DBLP:conf/cvpr/ZhangZCZZLSY23}
    & \std{0.416}{.006}    & \std{0.627}{.006}    & \std{0.745}{.006}    & \std{0.514}{.029}
    & \std{3.007}{.023}    & \std{10.92}{.108}    & \std[1]{1.570}{.039} \\
    & MoMask~\cite{DBLP:conf/cvpr/GuoMJW024}
    & \std[1]{0.433}{.007} & \std[1]{0.656}{.005} & \std[1]{0.781}{.005} & \std{0.204}{.011}
    & \std[1]{2.779}{.022} & -                    & \std{1.131}{.043}    \\
    \cline{2-9}
    & \bf{\OM-AR-2}  
    & \std{0.423}{.006}    & \std{0.633}{.005}    & \std{0.750}{0.006}   & \std{0.211}{.003} 
    & \std{3.048}{.032}    & \std{11.31}{.101} & \std{1.154}{.041} \\
    & \bf{\OM-DDM-2} 
    & \std[2]{0.427}{.006} & \std[2]{0.642}{.006} & \std{0.762}{.006}    & \std[1]{0.144}{.005}
    & \std{2.856}{.033}    & \std{10.92}{.163}    & \std[2]{1.325}{.051} \\
    & \bf{\OM-AR-4} 
    & \std{0.418}{.007}    & \std{0.619}{.007}    & \std{0.734}{.005}    & \std{0.297}{.002}
    & \std{3.075}{.036}    & \std[2]{11.15}{.085} & \std{1.102}{.039} \\
    & \bf{\OM-DDM-4} 
    & \std{0.421}{.007}    & \std{0.638}{.007}    & \std{0.760}{.006}    & \std{0.192}{.016}
    & \std{2.896}{.028}    & \std{10.81}{.115}    & \std{1.295}{.046} \\
    \bottomrule
  \end{tabular}
  }
  \vspace{-6 mm}
\end{table*}

%% file: tables/ablation_study.tex
\begin{table*}[t]
  \setlength\tabcolsep{4pt}
  \setlength\extrarowheight{1pt}
  \caption{\textbf{Ablation study of tokenizer configurations across DDM and AR planners.}
  Latent Dim is the codebook dimension, and Downrate denotes the temporal compression ratio (frames per latent).
  DiffusionHead (DH) applies independent, per-frame diffusion decoding using AdaIN-conditioned MLP blocks, while DiffusionConv (DC) augments it with residual 1D convolutions inserted between MLP blocks to model local temporal dependencies (kernel size $\in$ \{3, 5, 7, 9\}).
  ``DR'', ``Recon.'', ``Ctrl.'', and ``Err.'' stand for downsample rate, reconstruction, control, and error, respectively.
  \textcolor{red}{Red} and \textcolor{blue}{Blue} highlight the best and second-best results, respectively.}
  \vspace{-2 mm}
  \label{tab:quant_ablation}
  \resizebox{\textwidth}{!}{%
    \begin{tabular}{cccccc ccc ccc}
    \toprule
    \multirow{2}{*}{\#} &
    \multirow{2}{*}{\makecell{Latent\\Dim}} &
    \multirow{2}{*}{Decoder} &
    \multirow{2}{*}{DR} &
    \multirow{2}{*}{\makecell{Kernel\\Size}} &
    \multirow{2}{*}{\makecell{Recon.\\FID \darr}} &
    \multicolumn{3}{c}{DDM Planner} &
    \multicolumn{3}{c}{AR Planner} \\
    \cmidrule(lr){7-9}\cmidrule(lr){10-12}
    & & & & & &
    T2M FID \darr    &  Ctrl. FID \darr &  Ctrl. Err. \darr &
    T2M FID \darr    &  Ctrl. FID \darr &  Ctrl. Err. \darr \\
    \midrule
    1 & \multirow{2}{*}{768} & Conv & \multirow{2}{*}{4} & 3 &
    0.0704           & 
    0.0640           & 0.0601           & 0.2027               & 
    0.0935           & 0.0874           & 0.2604               \\ 
    2 & & DH & & N/A &
    0.0396           & 
    0.0678           & 0.0373           & 0.0051               & 
    0.0686           & 0.0739           & 0.0053               \\ 
    \midrule
    3 & 768 & \multirow{6}{*}{DC} & \multirow{6}{*}{4} & 3 & 
    0.0259           & 
    0.0500           & 0.0379           & \mean[1]{0.0047}     & 
    0.0671           & 0.0469           & \mean[1]{0.0049}     \\ 
    4 & 768 & & & 5 & 
    0.0244           & 
    \mean[2]{0.0394} & \mean[2]{0.0292} & \mean[2]{0.0049}     & 
    \mean[2]{0.0535} & \mean[2]{0.0461} & 0.0051               \\ 
    5 & 768 & & & 7 & 
    0.0288           & 
    0.0458           & \mean[1]{0.0279} & \mean[1]{0.0047}     & 
    0.0690           & \mean[1]{0.0454}           & \mean[2]{0.0050}     \\ 
    6 & 768 & & & 9 & 
    0.0400           & 
    0.0875           & 0.0400           & 0.0051               & 
    0.0720           & 0.0590           & 0.0054               \\ 
    7 & 512 & & & 5 & 
    \mean[1]{0.0177} & 
    0.0505           & 0.0457           & \mean[2]{0.0049}     & 
    0.0780           & 0.0649           & 0.0052               \\ 
    8 & 384 & & & 5 & 
    0.0292           & 
    0.0669           & 0.0428           & \mean[2]{0.0049}     & 
    0.0581           & 0.0692           & 0.0051              \\ 
    \midrule
    9 & \multirow{8}{*}{768} & \multirow{8}{*}{DC} & \multirow{2}{*}{1} & 5 & 
    0.0207           & 
    0.1849           & 0.0734           & 0.0050           & 
    0.1379           & 0.0630           & 0.0054           \\ 
    10 & & & & 7 & 
    \mean[2]{0.0188} & 
    0.1551           & 0.0542           & 0.0052           & 
    0.1465           & 0.0597           & 0.0056           \\ 
    \cline{4-12}
    11 & & & \multirow{2}{*}{2} & 5 & 
    0.0240           & 
    0.0459           & 0.0395           & 0.0050           & 
    0.0773           & 0.0474           & 0.0052           \\ 
    12 & & & & 7 & 
    0.0228           & 
    \mean[1]{0.0332} & 0.0353           & \mean[2]{0.0049}           & 
    0.0617           & 0.0468 & 0.0052           \\ 
    \cline{4-12}
    13 & & & \multirow{2}{*}{8} & 5 & 
    0.0429           & 
    0.0707           & 0.0304           & 0.0051           & 
    0.0682           & 0.0527           & 0.0052           \\ 
    14 & & & & 7 & 
    0.0492           & 
    0.0816           & 0.0362           & 0.0054           & 
    \mean[1]{0.0533} & 0.0708           & 0.0056           \\ 
    \cline{4-12}
    15 & & & \multirow{2}{*}{16} & 5 & 
    0.0600           & 
    0.0875           & 0.0474           & 0.0054           & 
    0.0788           & 0.1043           & 0.0054           \\ 
    16 & & & & 7 & 
    0.0894           & 
    0.1748           & 0.0818           & 0.0054           & 
    0.1580           & 0.1486           & 0.0055           \\ 
    \bottomrule
    \end{tabular}
  }
  \vspace{-6 mm}
\end{table*}

%% file: tables/ablation_low_ctrl.tex
\begin{table*}[t]
  \scriptsize
  \caption{\textbf{Effect of low-level control injection location.} 
  Comparison of injecting low-level control signals in the generator (token-space planning), the tokenizer decoder (diffusion decoding), evaluated with DDM and AR planners under ``Pelvis'' setting.}
  \vspace{-2 mm}
  \label{tab:control_injection_ablation}
  \begin{tabularx}{\linewidth}{Yc YY YY}
    \toprule
    \multicolumn{2}{c}{Low-level Condition} &
    \multicolumn{2}{c}{DDM Planner} &
    \multicolumn{2}{c}{AR Planner} \\
    \cmidrule(lr){1-2} \cmidrule(lr){3-4} \cmidrule(lr){5-6}
    Generator       & Tok. Decoder     &
    Ctrl. FID \darr & Ctrl. Err. \darr &
    Ctrl. FID \darr & Ctrl. Err. \darr \\
    \midrule
    \cmark          & \cmark           & 
    \mean[2]{0.029} & \mean[1]{0.0049} &
    \mean[2]{0.046} & \mean[1]{0.0051} \\
    \cmark          & \xmark           &
    \mean[1]{0.028} & 0.2170           &
    \mean[1]{0.037} & 0.2770           \\
    \xmark          & \cmark           &
    0.365           & \mean[2]{0.0056} &
    0.447           & \mean[2]{0.0063} \\
    \xmark          & \xmark           &
    0.039           & 0.6239           &
    0.053           & 0.6009           \\
    \bottomrule
  \end{tabularx}
  \vspace{-6 mm}
\end{table*}

%% file: sections/05_conclusion.tex
\section{Conclusion}

In this work, we bridge the strengths of continuous diffusion models for kinematic control and discrete token-based generators for semantic conditioning.
We introduce a three-stage \textbf{Perception--Planning--Control} paradigm that encodes conditions, plans discrete motion tokens, and synthesizes motion via diffusion-based decoding with fine-grained constraint enforcement. 
Within this framework, we proposed \OM, a diffusion-based discrete motion tokenizer that offloads low-level reconstruction to diffusion decoding, enabling compact and semantically informative tokens. Experiments on HumanML3D and KIT-ML demonstrate strong performance under aggressive token compression and significant gains on text-and-trajectory controllable generation, reducing trajectory error from 0.72 cm to 0.08 cm while using only one-sixth of the tokens.

%% file: sections/06_acknowledgement.tex
\paragraph{Acknowledgements}
This study is supported by the Ministry of Education, Singapore, under its MOE AcRF Tier 2 (MOET2EP20221-0012, MOE-T2EP20223-0002), and under the RIE2020 Industry Alignment Fund – Industry Collaboration Projects (IAF-ICP) Funding Initiative, as well as cash and in-kind contribution from the industry partner(s).

%% file: sections/99_appendix.tex
\section{From Semantic Usability to Detail Recoverability: Why MoTok Bridges Both}

In the main paper, we show that MoTok substantially improves downstream performance over existing tokenizers. A key reason is that MoTok upgrades the conventional “Decode” stage into a “Control” stage that can incorporate fine-grained kinematic conditions. This division of labor allows the planning stage to focus more on high-level conditions, leading to better overall generation.

In this section, we reveal another important reason: \textbf{\OM serves as an effective bridge between the motion space and the semantic space}. On the one hand, it provides stronger recovery of high-frequency motion details (Section~\ref{sec:better_decoding}). On the other hand, it preserves richer low-frequency motion characteristics that facilitate mapping textual semantics to discrete motion tokens (Section~\ref{sec:better_semantic}).

\subsection{Better low-level detail recoverability with diffusion decoding}
\label{sec:better_decoding}

To isolate the effect of the decoder on fine-grained motion reconstruction, we design a two-stage training protocol on HumanML3D and compare a standard VQ-VAE tokenizer against our MoTok tokenizer. In Stage-1, we train the tokenizer end-to-end to learn an encoder and a discrete codebook. In Stage-2, we freeze the Stage-1 encoder and codebook so that all variants share the same discrete tokens, and train the decoder only. We further consider a 2×2 design that combines the decoder architecture used in Stage-1 and Stage-2 (Conv vs. Diffusion), which enables a controlled and fair evaluation of reconstruction capacity under identical token information. We report reconstruction quality (Recon FID), downstream text-to-motion performance (T2M FID), and motion-to-text metrics (R@1/R@3/BLEU@1/BLEU@4) to reflect both low-level fidelity and semantic usability of the learned discrete representations.

Table~\ref{tab:two_stage_decoder_ablation} shows that, with the encoder and codebook frozen in Stage-2, replacing a conventional convolutional decoder with MoTok’s diffusion-based decoder consistently improves both reconstruction and downstream performance. In particular, the diffusion decoder achieves lower Recon FID, indicating stronger recovery of fine-grained details given the same discrete token sequence. This protocol provides a more principled and fair comparison than end-to-end training alone, since the gains cannot be attributed to increased token information capacity. The benefits also translate to text-to-motion generation, where diffusion-based decoding yields consistently lower T2M FID. Overall, these results demonstrate that \textbf{MoTok recovers motion details more faithfully under the same token information}.

\input{tables/ablation_decoder}

\subsection{Better semantic usability of discrete tokens}
\label{sec:better_semantic}

To compare the semantic information carried by different tokenizers, we further evaluate motion-to-text (M2T) captioning on top of MotionGPT. Interestingly, although M2T does not involve motion generation or reconstruction, the captioning model trained with tokens produced by a standard VQ-VAE in Stage-1 consistently underperforms the one trained with MoTok tokens across nearly all metrics. This indicates that MoTok tokens encode richer and more usable semantic information, which improves text–motion alignment and effectively reduces the burden on the downstream planner.

A plausible explanation is that the stronger diffusion-based reconstruction shifts high-frequency detail recovery to the decoder, \textbf{allowing MoTok's discrete tokens to devote less capacity to fine-grained kinematic variations and instead preserve more low-frequency information that captures overall motion structure and trends}. 

As a result, MoTok serves as an effective bridge: it recovers high-frequency motion details more faithfully under the same token information, while simultaneously simplifying semantic alignment in the planning stage, thereby connecting semantic and kinematic constraints more efficiently.

\section{From Richer Conditions to Broader Tasks: Generality of the Unified Paradigm}

\input{tables/ablation_cond}

For different types of conditions, we define two forms in the main paper: global and local. Specifically, we treat text features as global conditions and kinematic signals as local conditions. In this section, we present additional variants and usages of conditions to further demonstrate the generality and flexibility of our proposed conditioning framework.

\subsection{Inject Semantic Condition via Cross Attention}

For text, which serves as a global condition, we adopt an in-context-learning-style training strategy in the main paper. Another common practice in the motion generation literature is to inject text features through cross-attention. Here, we compare against this alternative design. As shown in Table~\ref{tab:ablation_cond}, compared with the in-context variant, the cross-attention version yields a clear improvement in R-Precision, indicating stronger text-motion alignment. However, this gain comes at the cost of a noticeable drop in FID. This suggests that different conditioning injection strategies lead to different trade-offs between R-Precision and FID.

\subsection{Global vs Local Hint Injection}

For kinematic conditions, we also explore injecting them in a global form. Specifically, we use a motion Transformer to extract global features from the hint sequence, and then place these features after the text features for in-context sequence modeling. The results are reported in Table~\ref{tab:ablation_cond} under the row “w/ Global Hint”. Note that we still retain the original local injection pathway in this setting. Empirically, adding global kinematic conditions brings almost no improvement across the evaluation metrics. This suggests that, for the current task, local conditioning alone is already sufficient, and introducing additional global kinematic guidance is unnecessary.

\subsection{Motion Editing as a new condition-driven task}

Another important application of kinematic conditions is motion editing. In this task, the input typically consists of a source motion together with a text prompt describing the desired modification. We inject the source motion as a local condition, in a manner similar to trajectory control. Specifically, we adopt a motion encoder architecture similar to that in \OM to extract features from the source motion, which are then used as position-wise embeddings. The text features are injected in the same way as in our other tasks. We provide several examples of this application setting in the demo video, which is trained under MotionFix~\cite{DBLP:conf/siggrapha/AthanasiouCDBV24} dataset.

\section{Additional Analyses on Downstream Behaviors}

\paragraph{Efficiency Comparison}
We evaluate our method on H100. It costs 2.63s to generate a single sequence with both textual prompt and trajectory conditions. As for MaskControl, the full version costs 32.79s in the same environment.

\input{tables/ablation_cfg_scale}

\paragraph{Ablation on CFG scale}
The CFG-scale ablation exhibits two consistent insights. First, nearly all variants follow a clear unimodal pattern: performance improves as the CFG scale increases from a small value, reaches an optimum, and then degrades when the scale becomes overly large. This indicates that moderate guidance best balances semantic alignment and motion realism, while excessive guidance can over-constrain sampling and harm fidelity. Second, the optimal CFG scale shifts with the temporal compression rate. As compression decreases (i.e., more tokens), even under the same masking and replacement strategy during training, the absolute token count increases and thus injects more motion-side information into the planning process. Consequently, the generator tends to rely more on motion tokens and time-synchronized cues, weakening the relative influence of the text condition. To compensate for this imbalance, inference requires a larger CFG scale to amplify the contribution of the text embedding and enforce stronger adherence to the given prompt, especially for low-downrate (high-token) settings.




\section{Implementation Details}

\subsection{Evaluation Metrics}

In this work, we mainly focus on three types of metrics:
\begin{enumerate}
    \item \textbf{Text-to-Motion (T2M)}: We evaluate both motion realism and text--motion alignment. Motion realism is measured by Fréchet Inception Distance (FID). Text--motion consistency is assessed by R-Precision (R@1/2/3) and Multimodal Distance (MM Dist). We further report Diversity (Div) to quantify variation across generated motions, and Multi-Modality (MM) to measure the ability to generate multiple plausible motions for the same text.
    \item \textbf{Motion-to-Text (M2T)}: We adopt standard captioning metrics, including BLEU@1/4, ROUGE-L, CIDEr, and BERTScore. In addition, we report R-Precision to measure alignment between generated captions and their corresponding motions. Below we define each metric.
    \item \textbf{Controllable Motion Generation}: We additionally evaluate \textbf{spatial-level control accuracy}, including Foot skating ratio, Trajectory error, Location error, and Average keyframe error. These metrics directly quantify whether generated motions satisfy the prescribed spatial control signals. 
\end{enumerate}

\textbf{R-Precision}
measures retrieval accuracy by computing the fraction of relevant items among the top-$R$ retrieved candidates:
\begin{equation}
R\text{-Prec} = \frac{| \text{Rel} \cap \text{Top-}R |}{R},
\end{equation}
where $\text{Rel}$ denotes the set of ground-truth matches and $\text{Top-}R$ denotes the retrieved set at rank $R$.

\textbf{Fr'echet Inception Distance (FID)}
measures the distance between real and generated feature distributions:
\begin{equation}
\text{FID} = | \mu_r - \mu_g |_2^2 + \mathrm{Tr}\big(\Sigma_r + \Sigma_g - 2(\Sigma_r \Sigma_g)^{1/2}\big),
\end{equation}
where $(\mu_r,\Sigma_r)$ and $(\mu_g,\Sigma_g)$ are the mean and covariance of real and generated features, respectively.

\textbf{Diversity}
quantifies sample-level variability by averaging pairwise distances:
\begin{equation}
\text{Div} = \frac{2}{M(M-1)} \sum_{i<j} | f(x_i) - f(x_j) |_2.
\end{equation}

\textbf{Multi-Modality}
measures conditional diversity among $K$ samples generated for the same text:
\begin{equation}
\text{MM} = \frac{2}{K(K-1)} \sum_{i<j} | f(x_i^t) - f(x_j^t) |_2.
\end{equation}

\textbf{Multimodal Distance}
evaluates cross-modal alignment in a shared embedding space:
\begin{equation}
\text{MM Dist} = \frac{1}{N} \sum_{i=1}^N d(f_{\text{text}}(t_i), f_{\text{motion}}(m_i)).
\end{equation}

\textbf{BLEU}
computes n-gram precision with a brevity penalty:
\begin{equation}
\text{BLEU@N} = \text{BP} \cdot \exp\left(\frac{1}{N} \sum_{n=1}^{N} \log p_{n}\right),
\end{equation}
with modified precision
\begin{equation}
p_{n} =
\frac{\sum_{\text{ngram} \in C} \min \Big(\text{Count}{C}(\text{ngram}), \max{R \in \text{Refs}} \text{Count}{R}(\text{ngram}) \Big)}
{\sum{\text{ngram} \in C} \text{Count}_{C}(\text{ngram})},
\end{equation}
and brevity penalty
\begin{equation}
\text{BP} =
\begin{cases}
1 & \text{if } c > r, \\
e^{(1-r/c)} & \text{if } c \leq r.
\end{cases}
\end{equation}

\textbf{ROUGE-L}
uses the longest common subsequence (LCS) and an $F$-measure:
\begin{equation}
\text{ROUGE-L} = \frac{(1+\beta^2) \cdot P_{LCS} \cdot R_{LCS}}{R_{LCS} + \beta^2 \cdot P_{LCS}}.
\end{equation}

\textbf{CIDEr}
computes TF-IDF weighted n-gram similarity:
\begin{equation}
\text{CIDEr}(c, S) = \frac{1}{|S|} \sum_{s \in S} \frac{g(c) \cdot g(s)}{|g(c)| |g(s)|}.
\end{equation}

\textbf{BERTScore}
measures semantic similarity via contextual embeddings:
\begin{equation}
\text{BERTScore}(c, r) = \frac{1}{|c|} \sum_{x \in c} \max_{y \in r} \cos\big(f(x), f(y)\big).
\end{equation}

\textbf{Spatial-level Metrics.}
Let ${\mathbf{p}_k,k\in\mathcal{K}}$ and ${\hat{\mathbf{p}}_k,k\in\mathcal{K}}$ denote the ground-truth and generated 3D positions at a set of keyframes $\mathcal{K}$ for the controlled joint (e.g., pelvis). We define the keyframe location error at keyframe $k$ as
\begin{equation}
e_k = \left| \hat{\mathbf{p}}_k - \mathbf{p}_k \right|_2.
\end{equation}
Given a threshold $\tau$ (e.g., 50,cm), we compute:

\textbf{(1) Average error.}
The mean keyframe position error is
\begin{equation}
\text{AvgErr} = \frac{1}{|\mathcal{K}|}\sum_{k\in\mathcal{K}} e_k.
\end{equation}

\textbf{(2) Location error.}
The fraction of keyframes whose error exceeds the threshold is
\begin{equation}
\text{LocErr} = \frac{1}{|\mathcal{K}|}\sum_{k\in\mathcal{K}} \mathbb{I}\big[e_k > \tau\big],
\end{equation}
where $\mathbb{I}[\cdot]$ is the indicator function.

\textbf{(3) Trajectory error.}
We mark a trajectory as unsuccessful if any keyframe violates the threshold, and report the failure ratio over $N$ test samples:
\begin{equation}
\text{TrajErr} = \frac{1}{N}\sum_{n=1}^{N}\mathbb{I}\Big[\max_{k\in\mathcal{K}} e^{(n)}_k > \tau\Big].
\end{equation}

\textbf{(4) Foot skating ratio.}
Let $\mathcal{F}$ denote the set of frames where the foot is in contact (estimated by a velocity/height heuristic), and $\hat{\mathbf{v}}t$ be the generated foot horizontal velocity at frame $t$. Foot skating ratio is computed as
\begin{equation}
\text{Skate} = \frac{1}{|\mathcal{F}|}\sum{t\in\mathcal{F}} \mathbb{I}\big[|\hat{\mathbf{v}}_t|_2 > \epsilon\big],
\end{equation}
where $\epsilon$ is a small threshold for detecting undesired sliding during contact.

\subsection{Details for Controllable Motion Generation}
\label{sec:impl_cmg}

\paragraph{Training Strategies}
For both the spatial control and text-to-motion (T2M) tasks, we adopt the same training protocol. Specifically, for DDM, AR, and MoTok, all models are trained for 24 epochs, with each epoch consisting of approximately 2,000 iterations.

For MoTok, training is conducted on 8 GPUs with a batch size of 512 per GPU. For DDM and AR, we also use 8 GPUs, but with a batch size of 64 per GPU. All models are optimized using AdamW, with an initial learning rate of 2e-4, which is decayed to 2e-5 at the 20th epoch.

\paragraph{Inference Strategies}
During inference, MoTok adopts a spaced diffusion scheme and uses Fast27, provided by GLIDE~\cite{DBLP:conf/icml/NicholDRSMMSC22}, as the sampling strategy, thereby substantially reducing the original 1,000 diffusion steps used in training. The inference speed could be further improved in the future by incorporating consistency models, which we leave for future work. For DDM, inference is performed with 10 autoregressive denoising steps. In contrast, AR predicts the output one token at a time, generating the token sequence sequentially.

\subsection{Details for Motion-to-Text}
\label{sec:impl_m2t}

\paragraph{Task Formulation}
For motion-to-text (M2T), the goal is to generate a natural-language description that faithfully summarizes a given motion sequence. Formally, given an input motion $\mathbf{x}={x_t}{t=1}^{T}$, where $x_t$ denotes the pose representation at time step $t$, the model predicts a caption $y={y_i}{i=1}^{L}$ consisting of $L$ tokens. We model M2T as conditional language generation by maximizing the likelihood $p(y \mid \mathbf{x})$. In our framework, the motion sequence is first encoded by the MoTok tokenizer into a discrete token sequence $\mathbf{z}={z_n}{n=1}^{N}$, where $N$ depends on the temporal downsampling rate. An autoregressive language decoder then generates the caption conditioned on $\mathbf{z}$:
\begin{equation}
p(y \mid \mathbf{x}) = p(y \mid \mathbf{z}) = \prod{i=1}^{L} p(y_i \mid y_{<i}, \mathbf{z}).
\end{equation}
This formulation allows the captioner to operate in a compact token space while leveraging MoTok to preserve semantically relevant motion content.

\paragraph{Implementation Details}
We implement the M2T captioner as a standard autoregressive transformer decoder conditioned on MoTok tokens. Specifically, we first encode motions into discrete tokens using the pretrained MoTok tokenizer (encoder + codebook), and keep the tokenizer fixed during captioning training. Token embeddings are obtained by table lookup and optionally augmented with positional embeddings. The captioning model takes the MoTok token sequence as conditioning context and generates text using teacher forcing during training, optimizing the cross-entropy loss over the ground-truth caption tokens. At inference time, we decode captions autoregressively using greedy decoding or beam search. Unless otherwise specified, we follow common M2T training practices in prior work, including using the HumanML3D captions as supervision, truncating/padding motion sequences to a maximum length, and applying standard text preprocessing (tokenization and vocabulary construction).

\input{tables/quant_m2t}

\paragraph{Full Quantitative Results}
As shown in Table~\ref{tab:quant_m2t}, replacing a conventional VQ-based motion tokenizer with \textbf{MoTok} consistently improves M2T captioning performance. 
This improvement suggests that MoTok produces discrete tokens that better capture motion semantics, enabling the language decoder to infer high-level action descriptions more effectively from the token sequence. 
By offloading fine-grained reconstruction to diffusion-based decoding, the discrete representation is freed from low-level details and can focus on semantic content. 
Consequently, the learned token space exhibits stronger semantic separability and serves as a more effective interface for text generation, supporting our design choice of using latent dimensionality as a semantically meaningful bottleneck rather than a reconstruction-driven representation.

%% file: tables/ablation_decoder.tex
\begin{table*}[!t]
\centering
\setlength\tabcolsep{5pt}
\setlength\extrarowheight{1pt}
\caption{\textbf{Two-stage training ablation of tokenizer decoders on HumanML3D.}
Stage-1 trains the tokenizer end-to-end. Stage-2 freezes the encoder and codebook and trains the decoder only.
All variants share the same discrete tokens in Stage-2, enabling a fair comparison of reconstruction capability under identical token information.}
\label{tab:two_stage_decoder_ablation}
\resizebox{\linewidth}{!}{%
\begin{tabular}{ll|cc|cccc}
\toprule
\multirow{2}{*}{\textbf{Stage-1 Decoder}} &
\multirow{2}{*}{\textbf{Stage-2 Decoder}} &
\multicolumn{1}{c}{\textbf{Reconstruction}} &
\multicolumn{1}{c|}{\textbf{T2M}} &
\multicolumn{4}{c}{\textbf{M2T}} \\
\cline{3-8}
& &
\textbf{FID} $\downarrow$ &
\textbf{FID} $\downarrow$ &
\textbf{R@1} $\uparrow$ &
\textbf{R@3} $\uparrow$ &
\textbf{BLEU@1} $\uparrow$ &
\textbf{BLEU@4} $\uparrow$ \\
\midrule
Conv (VQ)        & Conv (VQ)        & 0.0954 & 0.1271 & \multirow{2}{*}{\mean[2]{0.473}} & \multirow{2}{*}{\mean[2]{0.732}} & \multirow{2}{*}{\mean[2]{50.1}} & \multirow{2}{*}{\mean[2]{14.6}} \\
Conv (VQ)        & Diffusion (MoTok) & \mean[2]{0.0544} & 0.1098 &  &  &  &  \\
\hline
Diffusion (MoTok)& Conv (VQ)        & 0.1347 & \mean[2]{0.0718} & \multirow{2}{*}{\mean[1]{0.488}} & \multirow{2}{*}{\mean[1]{0.743}} & \multirow{2}{*}{\mean[1]{51.3}} & \multirow{2}{*}{\mean[1]{15.5}} \\
Diffusion (MoTok)& Diffusion (MoTok) & \mean[1]{0.0433} & \mean[1]{0.0642} &  &  &  &  \\
\bottomrule
\end{tabular}%
}
\end{table*}

%% file: tables/ablation_cond.tex
\begin{table*}[!h]
  \setlength\tabcolsep{4pt}
  \setlength\extrarowheight{1pt}
  \caption{\textbf{Controllable motion generation quantitative results on HumanML3D test set.} 
  Note that ``Traj. Err.'', ``Loc. Err.'', and ``Avg. Err.'' denote ``Trajectory Error'', ``Localization Error'', and ``Average Error'', respectively.
  \uarr(\darr) indicates that the values are better if the metric is larger (smaller). `$\rightarrow$' means closer to real data is better.
  \textcolor{red}{Red} and \textcolor{blue}{Blue} highlight the best and second-best results, respectively.}
  \vspace{-2 mm}
  \label{tab:ablation_cond}
  \resizebox{\textwidth}{!}{%
    \begin{tabular}{clccccccc}
    \toprule
        & Method
        & FID \darr
        & \makecell{R-Precision \uarr\\(Top-3)}
        & Diversity \rarr
        & \makecell{Foot Skating\\Ratio \darr}
        & \makecell{Traj. Err.\darr\\(50cm)}
        & \makecell{Loc. Err. \darr\\(50cm)}
        & \makecell{Avg. Err. \darr\\(m)} \\ 
    \midrule
    & Real Motion 
    & 0.002           & 0.797           & 9.503           & 0.000
    & 0.0000         & 0.0000         & 0.0000 \\ 
    \hline
    \multirow{7}{*}{\rotatebox{90}{One}} 
    & OmniControl~\cite{DBLP:conf/iclr/XieJZSJ24} 
    & 0.310            & 0.693            & \mean[1]{9.502} & 0.0608
    & 0.0617           & 0.0107           & 0.0404 \\
    & InterControl~\cite{DBLP:conf/nips/000100L024}
    & 0.178            & 0.669            & \mean[2]{9.498} & 0.0968
    & \mean[2]{0.0403} & \mean[2]{0.0031} & 0.0741 \\
    & CrowdMoGen~\cite{DBLP:journals/ijcv/CaoGZXGL26} 
    & 0.147            & 0.781  & 9.461           & 0.0829
    & \mean[1]{0.0000} & \mean[1]{0.0000} & \mean[2]{0.0028} \\
    & MaskControl~\cite{DBLP:conf/iccv/Pinyoanuntapong25} 
    & 0.083            & \mean[1]{0.805}  & 9.395           & 0.0545 
    & \mean[1]{0.0000} & \mean[1]{0.0000} & 0.0072 \\
    \cline{2-9}
    & \bf{\OM-DDM-2} 
    & \mean[1]{0.025}  & 0.779            & 9.756           & 0.0436
    & \mean[1]{0.0000} & \mean[1]{0.0000} & \mean[1]{0.0008} \\
    & \bf{w/ Cross Att.} 
    & \mean[2]{0.027}  & \mean[2]{0.799}            & 9.582           & \mean[2]{0.0432}
    & \mean[1]{0.0000} & \mean[1]{0.0000} & \mean[1]{0.0008} \\
    & \bf{w/ Global Hint} 
    & 0.028  & 0.775            & 9.876           & \mean[1]{0.0427}
    & \mean[1]{0.0000} & \mean[1]{0.0000} & \mean[1]{0.0008} \\
    \midrule
    \multirow{5}{*}{\rotatebox{90}{Two}} 
    & InterControl~\cite{DBLP:conf/nips/000100L024} 
    & 0.184            & 0.670            & \mean[1]{9.410}           & 0.0948 
    & \mean[2]{0.0475} & \mean[2]{0.0030} & 0.0911 \\
    & CrowdMoGen~\cite{DBLP:journals/ijcv/CaoGZXGL26} 
    & 0.178            & \mean[2]{0.777}  & 9.149           & 0.0865 
    & \mean[1]{0.0000} & \mean[1]{0.0000} & 0.0027 \\  
    \cline{2-9}
    & \bf{\OM-DDM-2} 
    & \mean[1]{0.016}  & 0.775            & 9.711           & \mean[2]{0.0456}
    & \mean[1]{0.0000} & \mean[1]{0.0000} & \mean[2]{0.0006} \\ 
    & \bf{w/ Cross Att.} 
    & 0.025  & \mean[1]{0.795}            & 9.678           & \mean[1]{0.0444}
    & \mean[1]{0.0000} & \mean[1]{0.0000} & \mean[1]{0.0005} \\ 
    & \bf{w/ Global Hint}  
    & \mean[2]{0.021}  & 0.776            & \mean[2]{9.608}           & \mean[1]{0.0444}
    & \mean[1]{0.0000} & \mean[1]{0.0000} & \mean[1]{0.0005} \\ 
    \midrule
    \multirow{5}{*}{\rotatebox{90}{Three}} 
    & InterControl~\cite{DBLP:conf/nips/000100L024} 
    & 0.199            & 0.673            & 9.352           & 0.0930 
    & \mean[2]{0.0487} & \mean[2]{0.0026} & 0.0969 \\
    & CrowdMoGen~\cite{DBLP:journals/ijcv/CaoGZXGL26} 
    & 0.192            & 0.778            & 9.169           & 0.0871 
    & \mean[1]{0.0000} & \mean[1]{0.0000} & \mean[2]{0.0030} \\    
    \cline{2-9}
    & \bf{\OM-DDM-2} 
    & \mean[1]{0.014}  & 0.786  & \mean[1]{9.519} & 0.0480
    & \mean[1]{0.0000} & \mean[1]{0.0000} & \mean[1]{0.0007} \\ 
    & \bf{w/ Cross Att.} 
    & \mean[2]{0.017} & \mean[1]{0.799}  & \mean[2]{9.440} & \mean[1]{0.0477}
    & \mean[1]{0.0000} & \mean[1]{0.0000} & \mean[1]{0.0007} \\ 
    & \bf{w/ Global Hint} 
    & \mean[2]{0.017}  & \mean[2]{0.779}  & 9.676 & \mean[2]{0.0479}
    & \mean[1]{0.0000} & \mean[1]{0.0000} & \mean[1]{0.0007} \\ 
    \bottomrule
    \end{tabular}
  }
  \vspace{-4 mm}
\end{table*}

%% file: tables/ablation_cfg_scale.tex
\begin{table*}[t]
  \setlength\tabcolsep{4pt}
  \setlength\extrarowheight{1pt}
  \caption{\textbf{Ablation on CFG scale.} We vary the CFG scale from 1.6 to 3.6 with a step size of 0.2. Each entry reports the evaluation result (e.g., FID; lower is better).}
  \label{tab:cfg_scale_ablation}
  \resizebox{\linewidth}{!}{
    \begin{tabular}{l ccccccccccc}
    \toprule
    Variant &
    1.6         & 1.8         & 2.0         & 2.2         & 2.4         & 2.6         &
    2.8         & 3.0         & 3.2         & 3.4         & 3.6         \\
    \midrule
    MoTok-DDM-1 & 
    0.5551      & 0.4139      & 0.3251      & 0.3130      & 0.2669      & 0.2093      & 
    0.2519      & 0.1730      & 0.1756      & 0.1769      & \bf{0.1551} \\
    MoTok-DDM-2 & 
    0.2792      & 0.2077      & 0.1741      & 0.1373      & 0.1069      & 0.0835      & 
    0.0630      & 0.0519      & \bf{0.0332} & 0.0403      & 0.0349      \\
    MoTok-DDM-4 & 
    0.1430      & 0.1023      & 0.0753      & 0.0466      & 0.0396      & \bf{0.0394} & 
    0.0423      & 0.0653      & 0.0798      & 0.0984      & 0.1272      \\
    MoTok-DDM-8 &
    0.1031      & 0.0772      & \bf{0.0707} & 0.0887      & 0.0993      & 0.1149      & 
    0.1477      & 0.1607      & 0.1949      & 0.2173      & 0.2755      \\
    MoTok-DDM-16 & 
    0.1036      & \bf{0.0875} & 0.0975      & 0.1104      & 0.1148      & 0.1282      & 
    0.1335      & 0.1418      & 0.1550      & 0.1604      & 0.1878      \\
    \midrule
    MoTok-AR-2  & 
    0.0857      & 0.0852      & 0.0803      & 0.0799      & 0.0783      & 0.0710      & 
    0.0699      & 0.0696      & \bf{0.0617} & 0.0692      & 0.0710      \\
    MoTok-AR-4  & 
    0.0907& 0.0839& \bf{0.0535}& 0.0548& 0.0629& 0.0753& 
    0.0974& 0.1361& 0.2071& 0.2207& 0.2582\\
    MoTok-AR-8  & 
    0.0554& \bf{0.0533}& 0.0551& 0.0669& 0.1194& 0.1457& 
    0.2292& 0.2652& 0.3351& 0.4201& 0.5580\\
    MoTok-AR-16 & 
    \bf{0.1580}& 0.1598& 0.1842& 0.1730& 0.2512& 0.3116& 
    0.3257& 0.3330& 0.4119& 0.4831& 0.5364\\
    \bottomrule
    \end{tabular}
  }
\end{table*}

%% file: tables/quant_m2t.tex
\begin{table*}[t]
  \setlength\tabcolsep{4pt}
  \setlength\extrarowheight{1pt}
  \caption{\textbf{Quantitative results of Motion-to-Text (M2T) on HumanML3D.} Note that ``Sep.'' and ``Uni.'' denote ``Seperated Model'' and ``Unified Model'', respectively. \uarr indicates higher is better. \textcolor{red}{Red} and \textcolor{blue}{Blue} highlight the best and second-best results, respectively.}
  \vspace{-2 mm}
  \label{tab:quant_m2t}
  \resizebox{\linewidth}{!}{
  \begin{tabular}{clccccccc}
    \toprule
    & Method
    & R@1\uarr     & R@3\uarr   & BLEU@1\uarr    & BLEU@4\uarr 
    & ROUGE-L\uarr & CIDEr\uarr & BERTScore\uarr \\
    \midrule
    & Real Motion 
    & 0.523            & 0.828         & -            & -
    & -                & -             & - \\
    \midrule
    \multirow{3}{*}{\rotatebox{90}{Sep.}}
    & TM2T~\cite{DBLP:conf/eccv/GuoZWC22}
    & 0.516           & 0.823           & 48.9           & 7.0  
    & 38.1            & 16.8            & 32.2 \\
    & LaMP~\cite{DBLP:conf/iclr/Li0HQZ0SDDY25}
    & 0.547           & 0.831           & 47.8           & 13.0
    & 37.1            & 28.9            & - \\
    & MG-MotionLLM~\cite{DBLP:conf/cvpr/WuXSKRBQS25}
    & \mean[1]{0.592} & \mean[2]{0.866} & -              & 8.1
    & -               & -               & \mean[1]{36.7} \\
    \midrule
    \multirow{4}{*}{\rotatebox{90}{Uni.}}
    & MotionGPT~\cite{DBLP:conf/aaai/ZhangHLTLC00YO24}
    & 0.543           & 0.827           & 48.2           & 12.5
    & 37.4            & 29.2            & 32.4 \\
    & MotionGPT2~\cite{DBLP:journals/corr/abs-2410-21747} 
    & 0.558           & 0.838           & 48.7           & 13.8
    & 37.6            & 29.8            & 32.6 \\
    & MoTe~\cite{DBLP:journals/corr/abs-2411-19786}       
    & \mean[2]{0.577} & \mean[1]{0.871} & 46.7           & 11.2
    & 37.4            & 31.5            & 30.3 \\
    \cmidrule(l){2-9}
    & \bf{Baseline-VQ}
    & 0.473           & 0.732           & \mean[2]{50.1}  & \mean[2]{14.6}
    & \mean[2]{40.8}  & \mean[2]{34.3}  & 34.3 \\
    & \bf{Baseline-MoTok}
    & 0.488           & 0.743           & \mean[1]{51.3}  & \mean[1]{15.5}
    & \mean[1]{41.4}  & \mean[1]{35.3}  & \mean[2]{34.8} \\
    \bottomrule
  \end{tabular}
  }
\end{table*}

%% file: references.bib
@String(PAMI  = {IEEE TPAMI})

@String(IJCV  = {IJCV})

@String(CVPR  = {CVPR})

@String(ICCV  = {ICCV})

@String(ECCV  = {ECCV})

@String(ICML  =	{ICML})

@String(NIPS  = {NeurIPS})

@String(SIGGRAPH  = {SIGGRAPH})

@String(TOG   = {ACM TOG})

@String(ACMMM = {ACM MM})

@String(ICLR  = {ICLR})

@String(AAAI = {AAAI})

@String(CVPRW= {CVPRW})

@String(ThDV= {3DV})

@article{DBLP:journals/ivc/OrmoneitBHK05,
  author       = {Dirk Ormoneit and
                  Michael J. Black and
                  Trevor Hastie and
                  Hedvig Kjellstr{\"{o}}m},
  title        = {Representing cyclic human motion using functional analysis},
  journal      = {Image and Vision Computing},
  volume       = {23},
  number       = {14},
  pages        = {1264--1276},
  year         = {2005}
}

@article{DBLP:journals/tog/MinC12,
  author       = {Jianyuan Min and
                  Jinxiang Chai},
  title        = {Motion graphs++: a compact generative model for semantic motion analysis
                  and synthesis},
  journal      = TOG,
  volume       = {31},
  number       = {6},
  pages        = {153:1--153:12},
  year         = {2012}
}

@inproceedings{DBLP:conf/mm/GuoZWZSDG020,
  author       = {Chuan Guo and
                  Xinxin Zuo and
                  Sen Wang and
                  Shihao Zou and
                  Qingyao Sun and
                  Annan Deng and
                  Minglun Gong and
                  Li Cheng},
  title        = {{Action2Motion}: Conditioned Generation of 3{D} Human Motions},
  booktitle    = ACMMM,
  year         = {2020}
}

@inproceedings{DBLP:conf/iccv/PetrovichBV21,
  author       = {Mathis Petrovich and
                  Michael J. Black and
                  G{\"{u}}l Varol},
  title        = {Action-Conditioned 3{D} Human Motion Synthesis with Transformer {VAE}},
  booktitle    = ICCV,
  year         = {2021}
}

@inproceedings{DBLP:conf/eccv/CervantesSSS22,
  author       = {Pablo Cervantes and
                  Yusuke Sekikawa and
                  Ikuro Sato and
                  Koichi Shinoda},
  title        = {Implicit Neural Representations for Variable Length Human Motion Generation},
  booktitle    = ECCV,
  year         = {2022}
}

@inproceedings{DBLP:conf/cvpr/BarsoumKL18,
  author       = {Emad Barsoum and
                  John R. Kender and
                  Zicheng Liu},
  title        = {{HP-GAN:} Probabilistic 3{D} Human Motion Prediction via {GAN}},
  booktitle    = CVPRW,
  year         = {2018}
}

@article{DBLP:journals/tog/HarveyYNP20,
  author       = {F{\'{e}}lix G. Harvey and
                  Mike Yurick and
                  Derek Nowrouzezahrai and
                  Christopher J. Pal},
  title        = {Robust motion in-betweening},
  journal      = TOG,
  volume       = {39},
  number       = {4},
  pages        = {60},
  year         = {2020}
}

@inproceedings{DBLP:conf/icml/LuCZLZ0S24,
  author       = {Shunlin Lu and
                  Ling{-}Hao Chen and
                  Ailing Zeng and
                  Jing Lin and
                  Ruimao Zhang and
                  Lei Zhang and
                  Heung{-}Yeung Shum},
  title        = {{HumanTOMATO:} Text-aligned Whole-body Motion Generation},
  booktitle    = ICML,
  year         = {2024}
}

@inproceedings{DBLP:conf/iclr/TevetRGSCB23,
  author       = {Guy Tevet and
                  Sigal Raab and
                  Brian Gordon and
                  Yonatan Shafir and
                  Daniel Cohen{-}Or and
                  Amit Haim Bermano},
  title        = {Human Motion Diffusion Model},
  booktitle    = ICLR,
  year         = {2023}
}

@inproceedings{DBLP:conf/nips/0001HSD0DBH24,
  author       = {Weihao Yuan and
                  Yisheng He and
                  Weichao Shen and
                  Yuan Dong and
                  Xiaodong Gu and
                  Zilong Dong and
                  Liefeng Bo and
                  Qixing Huang},
  title        = {{MoGenTS:} Motion Generation based on Spatial-Temporal Joint Modeling},
  booktitle    = NIPS,
  year         = {2024}
}

@inproceedings{DBLP:conf/cvpr/ChenJLHFCY23,
  author       = {Xin Chen and
                  Biao Jiang and
                  Wen Liu and
                  Zilong Huang and
                  Bin Fu and
                  Tao Chen and
                  Gang Yu},
  title        = {Executing your Commands via Motion Diffusion in Latent Space},
  booktitle    = CVPR,
  year         = {2023}
}

@inproceedings{DBLP:conf/cvpr/ZhangZCZZLSY23,
  author       = {Jianrong Zhang and
                  Yangsong Zhang and
                  Xiaodong Cun and
                  Yong Zhang and
                  Hongwei Zhao and
                  Hongtao Lu and
                  Xi Shen and
                  Shan Ying},
  title        = {Generating Human Motion from Textual Descriptions with Discrete Representations},
  booktitle    = CVPR,
  year         = {2023}
}

@inproceedings{DBLP:conf/cvpr/GuoMJW024,
  author       = {Chuan Guo and
                  Yuxuan Mu and
                  Muhammad Gohar Javed and
                  Sen Wang and
                  Li Cheng},
  title        = {{MoMask}: Generative Masked Modeling of 3{D} Human Motions},
  booktitle    = CVPR,
  year         = {2024}
}

@inproceedings{DBLP:conf/3dim/AhujaM19,
  author       = {Chaitanya Ahuja and
                  Louis{-}Philippe Morency},
  title        = {{Language2Pose}: Natural Language Grounded Pose Forecasting},
  booktitle    = ThDV,
  year         = {2019}
}

@inproceedings{DBLP:conf/cvpr/Lu0LCDDD0Z25,
  author       = {Shunlin Lu and
                  Jingbo Wang and
                  Zeyu Lu and
                  Ling{-}Hao Chen and
                  Wenxun Dai and
                  Junting Dong and
                  Zhiyang Dou and
                  Bo Dai and
                  Ruimao Zhang},
  title        = {{ScaMo:} Exploring the Scaling Law in Autoregressive Motion Generation
                  Model},
  booktitle    = CVPR,
  year         = {2025}
}

@inproceedings{DBLP:conf/nips/GuoHWZ25,
  author       = {Chuan Guo and
                  Inwoo Hwang and
                  Jian Wang and
                  Bing Zhou},
  title        = {{SnapMoGen:} Human Motion Generation from Expressive Texts},
  booktitle    = NIPS,
  year         = {2025}
}

@inproceedings{DBLP:conf/eccv/PetrovichBV22,
  author       = {Mathis Petrovich and
                  Michael J. Black and
                  G{\"{u}}l Varol},
  title        = {{TEMOS:} Generating Diverse Human Motions from Textual Descriptions},
  booktitle    = ECCV,
  year         = {2022}
}

@inproceedings{DBLP:conf/cvpr/GuoZZ0JL022,
  author       = {Chuan Guo and
                  Shihao Zou and
                  Xinxin Zuo and
                  Sen Wang and
                  Wei Ji and
                  Xingyu Li and
                  Li Cheng},
  title        = {Generating Diverse and Natural 3{D} Human Motions from Text},
  booktitle    = CVPR,
  year         = {2022}
}

@inproceedings{DBLP:conf/eccv/TevetGHBC22,
  author       = {Guy Tevet and
                  Brian Gordon and
                  Amir Hertz and
                  Amit H. Bermano and
                  Daniel Cohen{-}Or},
  title        = {{MotionCLIP}: Exposing Human Motion Generation to {CLIP} Space},
  booktitle    = ECCV,
  year         = {2022}
}

@inproceedings{DBLP:conf/iccv/PetrovichBV23,
  author       = {Mathis Petrovich and
                  Michael J. Black and
                  G{\"{u}}l Varol},
  title        = {{TMR:} Text-to-Motion Retrieval Using Contrastive 3{D} Human Motion
                  Synthesis},
  booktitle    = ICCV,
  year         = {2023}
}

@inproceedings{DBLP:conf/nips/HoJA20,
  author       = {Jonathan Ho and
                  Ajay Jain and
                  Pieter Abbeel},
  title        = {Denoising Diffusion Probabilistic Models},
  booktitle    = NIPS,
  year         = {2020}
}

@inproceedings{DBLP:conf/cvpr/RombachBLEO22,
  author       = {Robin Rombach and
                  Andreas Blattmann and
                  Dominik Lorenz and
                  Patrick Esser and
                  Bj{\"{o}}rn Ommer},
  title        = {High-Resolution Image Synthesis with Latent Diffusion Models},
  booktitle    = CVPR,
  year         = {2022}
}

@inproceedings{DBLP:conf/iccv/ZhongHZX23,
  author       = {Chongyang Zhong and
                  Lei Hu and
                  Zihao Zhang and
                  Shihong Xia},
  title        = {{AttT2M}: Text-Driven Human Motion Generation with Multi-Perspective
                  Attention Mechanism},
  booktitle    = ICCV,
  year         = {2023}
}

@inproceedings{DBLP:conf/nips/JiangCLYYC23,
  author       = {Biao Jiang and
                  Xin Chen and
                  Wen Liu and
                  Jingyi Yu and
                  Gang Yu and
                  Tao Chen},
  title        = {{MotionGPT}: Human Motion as a Foreign Language},
  booktitle    = NIPS,
  year         = {2023}
}

@inproceedings{DBLP:conf/cvpr/Pinyoanuntapong24,
  author       = {Ekkasit Pinyoanuntapong and
                  Pu Wang and
                  Minwoo Lee and
                  Chen Chen},
  title        = {{MMM:} Generative Masked Motion Model},
  booktitle    = CVPR,
  year         = {2024}
}

@article{DBLP:journals/bigdata/PlappertMA16,
  author       = {Matthias Plappert and
                  Christian Mandery and
                  Tamim Asfour},
  title        = {The {KIT} Motion-Language Dataset},
  journal      = {Big Data},
  volume       = {4},
  number       = {4},
  pages        = {236--252},
  year         = {2016}
}

@inproceedings{DBLP:conf/eccv/DaiCWLDT24,
  author       = {Wenxun Dai and
                  Ling{-}Hao Chen and
                  Jingbo Wang and
                  Jinpeng Liu and
                  Bo Dai and
                  Yansong Tang},
  title        = {{MotionLCM}: Real-Time Controllable Motion Generation via Latent Consistency
                  Model},
  booktitle    = ECCV,
  year         = {2024}
}

@inproceedings{DBLP:conf/nips/LiTLDH24,
  author       = {Tianhong Li and
                  Yonglong Tian and
                  He Li and
                  Mingyang Deng and
                  Kaiming He},
  title        = {Autoregressive Image Generation without Vector Quantization},
  booktitle    = NIPS,
  year         = {2024}
}

@inproceedings{DBLP:conf/iccv/KarunratanakulP23,
  author       = {Korrawe Karunratanakul and
                  Konpat Preechakul and
                  Supasorn Suwajanakorn and
                  Siyu Tang},
  title        = {Guided Motion Diffusion for Controllable Human Motion Synthesis},
  booktitle    = ICCV,
  year         = {2023}
}

@inproceedings{DBLP:conf/iclr/XieJZSJ24,
  author       = {Yiming Xie and
                  Varun Jampani and
                  Lei Zhong and
                  Deqing Sun and
                  Huaizu Jiang},
  title        = {{OmniControl:} Control Any Joint at Any Time for Human Motion Generation},
  booktitle    = ICLR,
  year         = {2024}
}

@inproceedings{DBLP:conf/nips/000100L024,
  author       = {Zhenzhi Wang and
                  Jingbo Wang and
                  Yixuan Li and
                  Dahua Lin and
                  Bo Dai},
  title        = {{InterControl:} Zero-shot Human Interaction Generation by Controlling
                  Every Joint},
  booktitle    = NIPS,
  year         = {2024}
}

@inproceedings{DBLP:conf/iclr/ShafirTKB24,
  author       = {Yoni Shafir and
                  Guy Tevet and
                  Roy Kapon and
                  Amit Haim Bermano},
  title        = {Human Motion Diffusion as a Generative Prior},
  booktitle    = ICLR,
  year         = {2024}
}

@inproceedings{DBLP:conf/eccv/GuoZWC22,
  author       = {Chuan Guo and
                  Xinxin Zuo and
                  Sen Wang and
                  Li Cheng},
  title        = {{TM2T:} Stochastic and Tokenized Modeling for the Reciprocal Generation
                  of 3D Human Motions and Texts},
  booktitle    = ECCV,
  year         = {2022}
}

@article{DBLP:journals/pami/ZhangCPHGYL24,
  author       = {Mingyuan Zhang and
                  Zhongang Cai and
                  Liang Pan and
                  Fangzhou Hong and
                  Xinying Guo and
                  Lei Yang and
                  Ziwei Liu},
  title        = {{MotionDiffuse}: Text-Driven Human Motion Generation With Diffusion
                  Model},
  journal      = PAMI,
  volume       = {46},
  number       = {6},
  pages        = {4115--4128},
  year         = {2024}
}

@inproceedings{DBLP:conf/iccv/ZhangGPCHLYL23,
  author       = {Mingyuan Zhang and
                  Xinying Guo and
                  Liang Pan and
                  Zhongang Cai and
                  Fangzhou Hong and
                  Huirong Li and
                  Lei Yang and
                  Ziwei Liu},
  title        = {{ReMoDiffuse}: Retrieval-Augmented Motion Diffusion Model},
  booktitle    = ICCV,
  year         = {2023}
}

@article{DBLP:journals/ijcv/CaoGZXGL26,
  author       = {Yukang Cao and
                  Xinying Guo and
                  Mingyuan Zhang and
                  Haozhe Xie and
                  Chenyang Gu and
                  Ziwei Liu},
  title        = {{CrowdMoGen:} Zero-Shot Text-Driven Collective Motion Generation},
  journal      = IJCV,  
  volume       = {134},
  number       = {1},
  pages        = {29},
  year         = {2026}
}

@article{DBLP:preprint/arxiv/2505-05474,
  author       = {Beichen Wen and
                  Haozhe Xie and
                  Zhaoxi Chen and
                  Fangzhou Hong and
                  Ziwei Liu},
  title        = {3{D} Scene Generation: {A} Survey},
  journal      = {arXiv 2505.05474},
  year         = {2025}
}

@article{DBLP:journals/tog/HenterAB20,
  author       = {Gustav Eje Henter and
                  Simon Alexanderson and
                  Jonas Beskow},
  title        = {{MoGlow:} probabilistic and controllable motion synthesis using normalising
                  flows},
  journal      = TOG,
  volume       = {39},
  number       = {6},
  pages        = {236:1--236:14},
  year         = {2020}
}

@inproceedings{DBLP:conf/iccv/Pinyoanuntapong25,
  author       = {Ekkasit Pinyoanuntapong and
                  Muhammad Usama Saleem and
                  Korrawe Karunratanakul and
                  Pu Wang and
                  Hongfei Xue and
                  Chen Chen and
                  Chuan Guo and
                  Junli Cao and
                  Jian Ren and
                  Sergey Tulyakov},
  title        = {{MaskControl:} Spatio-Temporal Control for Masked Motion Synthesis
},
  booktitle    = ICCV,
  year         = {2025}
}

@inproceedings{DBLP:conf/icml/RadfordKHRGASAM21,
  author       = {Alec Radford and
                  Jong Wook Kim and
                  Chris Hallacy and
                  Aditya Ramesh and
                  Gabriel Goh and
                  Sandhini Agarwal and
                  Girish Sastry and
                  Amanda Askell and
                  Pamela Mishkin and
                  Jack Clark and
                  Gretchen Krueger and
                  Ilya Sutskever},
  title        = {Learning Transferable Visual Models From Natural Language Supervision},
  booktitle    = ICML,
  year         = {2021}
}

@article{DBLP:journals/corr/abs-2601-22153,
  title        = {{DynamicVLA:} A Vision-Language-Action Model for 
                  Dynamic Object Manipulation},
  author       = {Haozhe Xie and 
                  Beichen Wen and 
                  Jiarui Zheng and 
                  Zhaoxi Chen and 
                  Fangzhou Hong and 
                  Haiwen Diao and 
                  Ziwei Liu},
  journal      = {arXiv 2601.22153},
  year         = {2026}
}

@article{DBLP:journals/corr/abs-2510-26794,
  author       = {Jing Lin and
                  Ruisi Wang and
                  Junzhe Lu and
                  Ziqi Huang and
                  Guorui Song and
                  Ailing Zeng and
                  Xian Liu and
                  Chen Wei and
                  Wanqi Yin and
                  Qingping Sun and
                  Zhongang Cai and
                  Lei Yang and
                  Ziwei Liu},
  title        = {The Quest for Generalizable Motion Generation: Data, Model, and Evaluation},
  journal      = {arXiv 2510.26794},
  year         = {2025}
}

@inproceedings{DBLP:conf/iclr/Li0HQZ0SDDY25,
  author       = {Zhe Li and
                  Weihao Yuan and
                  Yisheng He and
                  Lingteng Qiu and
                  Shenhao Zhu and
                  Xiaodong Gu and
                  Weichao Shen and
                  Yuan Dong and
                  Zilong Dong and
                  Laurence Tianruo Yang},
  title        = {{LaMP:} Language-Motion Pretraining for Motion Generation, Retrieval,
                  and Captioning},
  booktitle    = ICLR,
  year         = {2025}
}

@inproceedings{DBLP:conf/cvpr/WuXSKRBQS25,
  author       = {Bizhu Wu and
                  Jinheng Xie and
                  Keming Shen and
                  Zhe Kong and
                  Jianfeng Ren and
                  Ruibin Bai and
                  Rong Qu and
                  Linlin Shen},
  title        = {{MG-MotionLLM:} {A} Unified Framework for Motion Comprehension and Generation
                  across Multiple Granularities},
  booktitle    = CVPR,
  year         = {2025}
}

@inproceedings{DBLP:conf/aaai/ZhangHLTLC00YO24,
  author       = {Yaqi Zhang and
                  Di Huang and
                  Bin Liu and
                  Shixiang Tang and
                  Yan Lu and
                  Lu Chen and
                  Lei Bai and
                  Qi Chu and
                  Nenghai Yu and
                  Wanli Ouyang},
  title        = {{MotionGPT:} Finetuned {LLMs} Are General-Purpose Motion Generators},
  booktitle    = AAAI,
  year         = {2024}
}

@article{DBLP:journals/corr/abs-2410-21747,
  author       = {Yuan Wang and
                  Di Huang and
                  Yaqi Zhang and
                  Wanli Ouyang and
                  Jile Jiao and
                  Xuetao Feng and
                  Yan Zhou and
                  Pengfei Wan and
                  Shixiang Tang and
                  Dan Xu},
  title        = {{MotionGPT-2:} {A} General-Purpose Motion-Language Model for Motion
                  Generation and Understanding},
  journal      = {arXiv 2410.21747},
  year         = {2024}
}

@article{DBLP:journals/corr/abs-2411-19786,
  author       = {Yiming Wu and
                  Wei Ji and
                  Kecheng Zheng and
                  Zicheng Wang and
                  Dong Xu},
  title        = {{MoTe:} Learning Motion-Text Diffusion Model for Multiple Generation
                  Tasks},
  journal      = {arXiv 2411.19786},
  year         = {2024}
}

@inproceedings{DBLP:conf/icml/NicholDRSMMSC22,
  author       = {Alexander Quinn Nichol and
                  Prafulla Dhariwal and
                  Aditya Ramesh and
                  Pranav Shyam and
                  Pamela Mishkin and
                  Bob McGrew and
                  Ilya Sutskever and
                  Mark Chen},
  title        = {{GLIDE:} Towards Photorealistic Image Generation and Editing with
                  Text-Guided Diffusion Models},
  booktitle    = ICML,
  year         = {2022}
}

@inproceedings{DBLP:conf/siggrapha/AthanasiouCDBV24,
  author       = {Nikos Athanasiou and
                  Alp{\'{a}}r Cseke and
                  Markos Diomataris and
                  Michael J. Black and
                  G{\"{u}}l Varol},
  title        = {{MotionFix:} Text-Driven 3{D} Human Motion Editing},
  booktitle    = SIGGRAPH,
  pages        = {44:1--44:11},
  year         = {2024}
}
